\documentclass[preprint,12pt]{elsarticle}

\usepackage[margin=2.5cm]{geometry}
\usepackage{amssymb}
\usepackage{lineno,hyperref}
\usepackage{graphicx}
\usepackage{amsmath}
\usepackage{multirow}
\usepackage[ruled, linesnumbered]{algorithm2e}
\usepackage{setspace}
\usepackage{algorithmic}
\usepackage{xcolor}
\usepackage{float}
\usepackage{graphicx}

\journal{Decision Support Systems}

\begin{document}

\begin{frontmatter}

\title{An Interpretable Probabilistic Approach for Demystifying Black-box Predictive Models}

%\title{Demystifying Predictive Black-Box Models: An Interpretable Probabilistic Approach}

\author{Catarina Moreira\corref{cor1}}
		\cortext[cor1]{Corresponding author}
		\ead{catarina.pintomoreira@qut.edu.au}
\author{Yu-Liang Chou}
\author{Mythreyi Velmurugan}
\author{Chun Ouyang} 
\author{Renuka Sindhgatta}
\author{Peter Bruza}
 
\address{School of Information Systems, Queensland University of Technology, Brisbane, Australia
\vspace*{-1.25\baselineskip}}

\begin{abstract}
%(Max. 250 words) 
The use of sophisticated machine learning models for critical decision making is faced with a challenge that these models are often applied as a `black-box’. This has led to an increased interest in interpretable machine learning, where \textit{post hoc} interpretation presents a useful mechanism for generating interpretations of complex learning models. In this paper, we propose a novel approach underpinned by an extended framework of Bayesian networks for generating post hoc interpretations of a black-box predictive model. The framework supports extracting a Bayesian network as an approximation of the black-box model for a specific prediction. Compared to the existing post hoc interpretation methods, the contribution of our approach is three-fold. Firstly, the extracted Bayesian network, as a probabilistic graphical model, can provide interpretations about not only what input features, but also why these features contributed to a prediction. Secondly, for complex decision problems with many features, a Markov blanket can be generated from the extracted Bayesian network to provide interpretations with a focused view on those input features that directly contributed to a prediction. Thirdly, the extracted Bayesian network enables the identification of four different rules which can inform the decision-maker about the confidence level in a prediction, thus helping the decision-maker assess the reliability of predictions learned by a black-box model. We implemented the proposed approach, applied it in the context of two well-known public datasets and analysed the results, which are made available in an open-source repository. 
%\vspace*{.25\baselineskip}
\end{abstract}

\begin{keyword}
Interpretable machine learning, post hoc interpretation, probabilistic inference, Bayesian Network, predictive analytics
\vspace*{-.25\baselineskip}
\end{keyword}

\end{frontmatter}

\section{Introduction}
\label{sec:intro}

%%%%%%%% THE CONTEXT: Black-box problem %%%%%%%%
The rapidly growing adoption of Artificial intelligence (AI) has led to the development of supervised machine learning, in particular, deep neural networks, for generating predictions of high accuracy~\cite{corr/abs-1901-04592}. While the advancement has the potential to make significant improvement to the state-of-the-art in operational decision making across various business domains and processes, the underlying models are often opaque and do not provide the decision-maker with any understanding of their internal predictive mechanisms. This opaqueness in machine learning models is known as the \textit{black-box} problem. 
Immediate consequences of trusting predictions from opaque models might result in severe losses for businesses (and people), unfair job losses, or even lead to negative impacts in certain societal groups (for instance, racial and gender discrimination)~\cite{corr/abs-2001-02478}. 
This has posed an open challenge to data scientists and business analysts on how to endow machine intelligence with capabilities to explain the underlying predictive mechanisms in a way that helps decision-makers understand and scrutinize the machine learned predictions.

%%%%%%%% BACKGROUND: Interpretable ML & post-hoc interpretation %%%%%%%%
The recent body of literature in machine learning has emphasised the need to interpret and explain the (machine) learned predictions. Methods and techniques have been proposed for explaining black-box models which are known as interpretable machine learning~\cite{guidotti2018} or, in a broader context, explainable AI (XAI)~\cite{lakkaraju2019}. So far, there exist two different mechanisms to address model interpretability. One is to have an interpretable model that provides transparency at three levels: the entire model, the individual components and the learning algorithm~\cite{lipton2018}. For example, both linear regression models and decision tree models are interpretable models. Another mechanism to address model interpretability is via \textit{post hoc} interpretation, in which case, explanations and visualisations are extracted from a learned model, that is, after the model has been trained, and as such they are model agnostic. This is particularly useful for generating model interpretations for those complex machine learning models (such as deep neural networks) that have low transparency and are hard to be transformed into an interpretable model (i.e., a `white-box') due to their sophisticated internal representations. The existing post hoc interpretation techniques (see a review in~\cite{guidotti2018}) present knowledge about the various levels of impact of individual input features on the corresponding prediction.
%``To interpret means to give or provide the meaning or to explain and present in understandable terms"~\cite{guidotti2018}. 

%%%%%%%% CONTRIBUTION: Our proposed framework %%%%%%%%
In this paper, we propose a novel approach underpinned by an extended framework of Bayesian networks for generating post hoc interpretations of a black-box predictive model, with a focus on providing interpretations for any instance of prediction learned by the model (known as local interpretations). We name this framework the \textit{Local Interpretation-Driven Abstract Bayesian Network} (LINDA-BN), which supports extracting a Bayesian network as an approximation (or an abstraction) of a black-box model for a specific prediction learned from any given input. We implemented our approach, applied it in the context of two well-known public datasets and analysed the results, which are made available in an open-source repository. % together with those obtained from two typical existing methods for post hoc interpretation.  
Compared to the existing post hoc interpretation methods, the contribution of our approach is three-fold. 
\begin{itemize}
    \item The extracted Bayesian network not only can provide interpretations about \textit{what} input features contributed to the corresponding prediction. As a probabilistic graphical model, it also represents knowledge about dependencies (in form of conditional probabilities) between input features and prediction, thus generating interpretations about \textit{why} certain input features contributed to the prediction. 
    \item For complex decision problems with a large number of features, the extracted Bayesian network is often complicated to be analysed by human. In this case, LINDA-BN supports generating a Markov blanket from the extracted Bayesian network. The Markov blanket determines the boundaries of a decision system in a statistical sense, and presents a graph structure covering a decision (e.g., a prediction), its parents, children, and the parents of the children. As such, the Markov blanket of the extracted Bayesian network provides interpretation with a focused view on those input features that directly contributed to the corresponding prediction. 
    \item The extracted Bayesian network enables the identification of four different rules which can inform the decision-maker about the confidence level in a given prediction. As such, the interpretations provided in our approach can help the decision-maker assess the reliability of predictions learned by a black-box model. 
\end{itemize}

%%%%%%%% Outline of Paper %%%%%%%%
In the rest of the paper, we continue to introduce the relevant concepts and review the related research efforts in Section~\ref{sec:backg}. 
We present our approach underpinned by the framework LINDA-BN in Section~\ref{sec:frame}.
Next, we report the experiments and discuss the results of analysis in Section~\ref{sec:eval}. 
Finally, we conclude the paper with an outlook to future work (Section 5).

\section{Background and Related Work}
\label{sec:backg}

In this section, we present the main concepts that are used throughout our work, and 
review research efforts that are related to the proposed framework.

\subsection{Concepts}

Prior to discussing existing work that relates to our approach on providing intepretations of a \textit{black box} machine learning model prediction, we note the following definitions:

\begin{itemize}
    \item \textbf{Black box predictor:} It is a machine learning opaque model, whose internals are either unknown to the observer or they are known but are not understandable by humans. 
    
    \item \textbf{Interpretability:} The ability to extract symbolic information out of a black box that can provide the meaning in understandable terms to a human \citep{doshivelez2017}.
  
     \item \textbf{Explainability:} The ability to highlight decision-relevant parts of the used representations of the algorithms and active parts in the algorithmic model, that either contribute to the model accuracy on the training set, or a specific prediction for one particular observation~\citep{Holzinger19}. %In our work, explainability  communicating in a form that can be understood by human when  decisions.

\end{itemize}

One can see interpretability as the extraction of symbolic information from the black box (machine-level) that already needs some degree of semantics, and explainability as the conversion of this symbolic information to a human understandable way (human-level). 

\subsection{Related Work}

Various approaches have been proposed in the literature to address the problem of interpretability. Generally, this problem can be classified into two major models: Interpretable models and model agnostic (post-hoc) models. %These approaches are summarized in Figure \ref{fig:Explainable_models}.
%\begin{figure}[h!]
%     \resizebox{\columnwidth}{!} {
%    \includegraphics{Explainable_models.pdf}
%    }
%    \caption{The different types of explainable models proposed in the literature}
%    \label{fig:Explainable_models}
%\end{figure}

Interpretable models are by design already interpretable, providing the decision-maker a transparent white box approach for prediction.  Decision tree, logistic regression, and linear regression are commonly used interpretable models. These models have been used to explain predictions of specific prediction problems~\cite{Siering2018}. Model-agnostic approaches,  on the other hand, refer to the deriving explanations from a black box predictor by extracting information about the underlying mechanisms of the system. In addition, studies have focused on providing model-specific post-hoc explanations~\cite{KIMDSS2020}. The focus of our work is to build model-agnostic post-hoc methods as they have flexibility of being applied to any predictive model as compared to model-specific post-hoc approaches . To discover the demystifying predictive black box models, we focus on the widely cited post-hoc models that include LIME~\cite{Ribeiro16}, SHAP~\cite{Lundberg17}, and Counterfactual explanation in this work.

\subsubsection{LIME}

Local Interpretable Model-agnostic Explanations (LIME)~\cite{Ribeiro16} explains the predictions of any classifier by approximating it with an locally faithful interpretable model. Hence, LIME generates local interpretations by perturbing a sample around the input vector within a local decision boundary \citep{Elshawi19,Ribeiro16}. Each feature is associated with a weight that is computed using a similarity function that measures the distances between the original instance prediction and the predictions of the sampled points in the local decision boundary. Linear regression is learned to determine the local importance of each feature.

LIME has been extensively applied in the literature. For instance, \citet{Stiffler18} used LIME to generate salience maps of a certain region showing which parts of the image affect how the black box model reaches a classification for a given test image. \citet{Tan17} apply LIME  to demonstrate the presence of uncertainty in the explanations that could raise concerns in the use of the black box model and diminish the value of the explanations.  different sources of uncertainty in the explanation. Their work demonstrates the presence of three sources of uncertainty: randomness in the sampling procedure, variation with sampling proximity, and variation in explained model across different data points. Anchor \citep{Ribeiro18} is an extension of LIME that attempts to address some of the limitations by maximizing likelihood on how a certain feature might contribute to a prediction. Anchor introduces IF-THEN rules as explanations as well as the notion of coverage, which allows the decision-maker to understand the boundaries in which the generated explanations are valid.

\subsubsection{SHAP}

The SHAP (SHapley Additive exPlanations) is an explanation method which uses Shapley values \cite{Shapley52} from coalitional game theory to fairly distribute the gain among players, where contributions of players are unequal~\citep{Lundberg17}. Shapely values are a concept in economics and game theory and consist in a method to fairly distribute the payout of a game among a set of players. One can map these game theoretic concepts directly to an XAI approach: a game is the prediction task for a single instance; the players are the feature values of the instance that collaborate to receive the gain. This gain consists of the difference between the Shapley value of the prediction and the average of the Shapley values of the predictions among the feature values of the instance to be explained \cite{Strumbelj13}. 

\citet{Strumbelj13} claim that in a coalition game, it is usually assumed that $n$ players form a grand coalition that has a certain value. Given that we  know how much each smaller (subset) coalition would have been worth, the goal is to distribute the value of the grand coalition among players fairly (that is, each player should receive a fair share, taking into account all sub-coalitions). \citet{Lundberg17} on the other hand, present an explanation using SHAP values and the differences between them to estimate the gains of each feature. 

In order to fairly distribute the payoff amongst players in a collaborative game, SHAP makes use of two fairness properties: (1) Additivity, which states that amounts must sum up to the final game result, and (2) Consistency, which states that if one player contributes more to the game, (s)he cannot get less reward. 

In terms of related literature, \citet{Ariza20} adopted SHAP values to assess logistic regression model and several machine learning algorithms for granting scoring in P2P (peer-to-peer) lending, the authors point out SHAP values can reflect dispersion, non-linearity and structural breaks in the relationships between each feature and the target variable. They concluded that the SHAP can provide accurate and transparent results on the credit scoring model. \citet{Parsa20} also highlight that SHAP could bring insightful meanings to interpret  prediction outcomes. For instance, one of the techniques in the model, XGBoost, not only is capable of evaluating the global importance of the impacts of features on the output of a model, but it can also extract complex and non-linear joint impacts of local features.

\subsubsection{Probabilistic graphical model}

The literature of interpretable methods for explainable AI based on probabilistic graphical models (PGM) is mostly dominated by models based on counterfactual reasoning in order to derive explanations for a scpecic local datapoint.

The counterfactual explanation based on PGM comprises of a conditional assertion whose antecedent is false and whose consequent describes how the world would have been if the antecedent had occurred. It provides interpretations as a mean to point out which changes would be necessary to accomplish the desired goal, rather than supporting the understanding of why the current situation had a certain predictive outcome \citep{Wachter18}. For instance, in a scenario where a machine learning algorithm assesses whether a person should be granted a loan or not, a counterfactual explanation of \textit{why} a person did not have a loan granted could be in a form of a scenario \textit{if your income was greater than $ \$15,000$ you would be granted a loan} \citep{Mothila20}. Unlike other explanation methods that depend on approximating an interpretable model within a perturbed decision boundary, counterfactual explanations have the strength that it is always truthful to the underlying model by providing direct outputs of the algorithms \citep{Ribeiro16}.

Counterfactual explanations are part of causal inference methods, which  are based on causal reasoning. and is focused on the estimation of the causal effects from treatments and actions \citep{Pear19}. In 2000, Pearl proposed a framework (the ladder of causation) that proposes different levels of causal relationships during causal inference. Level 1, \textit{Association}, entails the sensing of regularities or patterns in the input data, expressed as relations; it focuses on the question \textit{what}. Level 2, \textit{Intervention}, predicts the effects of deliberate actions, expressed as causal relationships. And Level 3, \textit{Counterfactuals}, involve constructing a theory of the world that explains why certain actions have specific effects and what happens is the absence of such actions \citep{Pear19}. A simple and naive approach for generating counterfactual explanations is searching by trial and error. In this approach the feature values are randomly changed for the instance of interest and stops searching when the desired output is predicted.

The notion of counterfactual model has been investigated by a few researchers. In 2013, the counterfactual approach has been proposed for the evolution of advertisement placement in search engines~\citep{Bottou13}. \citet{johansson2016learning} claim that the counterfactual thinking has been adopted in the context of machine learning applications to predict the result of several different actions, policies, and interventions using non-experiment data. Moreover, the Counterfactual Gaussian Process (CGP) approach has been created by \citet{schulam2017reliable} for modelling the effects of sequences of actions on continuous time series data and facilitate the reliability of medical decisions \citep{Neto20}.

Although counterfactual explanation are useful, they do not explain why a certain prediction is made. On the contrary, they assume a hypothetical scenario where the prediction would be contrary to the output of that particular data point. Our approach aims to use probabilistic model to provide local explanations that provide insights into the features influencing a datapoint.

\section{The Local Interpretation-Driven Abstract Bayesian Network Framework} 
\label{sec:frame}

In this section, we present our framework built upon an extended framework of Bayesian networks that can generate post hoc interpretations for a single data point of prediction: the local interpretation-driven abstract Bayesian network (LINDA). We start with a brief introduction to Bayesian networks (Section~\ref{subsec:bn}) and structure learning (Section~\ref{subsec:learning}). 
Readers that are familiar with the knowledge can proceed directly to the proposed framework (Sections~\ref{subsec:model} to~\ref{subsec:rules}).

%We name this framework the \textit{Local Interpretation-Driven Abstract Bayesian Network} (LINDA-BN). The reason why we refer to the extracted Bayesian network as \textit{abstract} is simply due to the fact that it is obtained as an \textit{approximation} (or an abstraction) of a black-box predictive model. % represented by the black-box predictor. Given an input vector $x$ and a black-box predictive model $\hat{y}$, the computation of LINDA-BN mainly consists of three steps: 
%i) to generate a set of permutations $x'$, 
%ii) to predict using the permuted points and record the predictions in $\hat{y}(x')$, and 
%iii) to learn a Bayesian network that represents the conditional dependencies between the features of $x'$ and predictions $\hat{y}(x')$. 

%This framework consists in three major steps: given a single input vector, $x$ and a black box model, $\hat{y}$, (1) generate a set of permutations, $x'$, (2) classify the permuted points, $\hat{y}(x')$, and (3) learn a Bayesian network structure that represents the conditional dependencies between the features of $x'$ and the predictions,  $\hat{y}(x')$.

%In the next sections, we introduce fundamental concepts regarding Bayesian networks (Section~\ref{subsec:bn}) and learning in Bayesian networks (Section~\ref{subsec:learning}). If the reader is already familiar with these concepts, we suggest the reader to go directly to our proposed framework (Section~\ref{subsec:model}).

\subsection{Bayesian Networks}
\label{subsec:bn}

A Bayesian Network (BN) is a directed acyclic graph in which each node represents a random variable, and each edge represents a direct influence from the source node to the target node. The graph represents (in)dependence relationships between variables, and each node is associated with a conditional probability table that specifies a distribution over the values of the node given each possible joint assignment of the  values of its parents~\citep{Pearl88}.

Bayesian networks can represent essentially any full joint probability distribution, which can be computed using the chain rule in probability theory~\citep{russel10}. %for Bayesian networks~\citep{russel10}. 
Let $\mathcal{G}$ be a BN graph over the variables $X_1, \cdots, X_n$. We say that a probability distribution, $Pr$, over the same space factorizes according to $\mathcal{G}$, if $Pr$ can be expressed using the following equation~\cite{koller09prob}: \vspace*{-.25\baselineskip}

\begin{equation}
Pr( X_1, \dots, X_n ) = \prod_{i=1}^n  Pr( X_i | Pa_{X_i} ).
\label{eq:joint}
\end{equation}

In Equation~\ref{eq:joint}, $Pa_{X_i}$ corresponds to all the parent variables of $X_i$. The graph structure of the network, together with the associated factorization of the joint distribution allows the probability distribution to be used effectively for inference (i.e. answering queries using the distribution as our model of the world). For some query $Y$ and some observed variable $e$, the exact inference in Bayesian networks is given by the following equation~\citep{koller09prob}: \vspace*{-\baselineskip}

\begin{equation}
Pr( Y | E = e ) = \alpha Pr(Y, e) = \alpha  \sum_{ w \in W } Pr(Y, e, w),  \text{~~~~~with~} \alpha = \frac{1}{\sum_{y \in Y} Pr(y, e).}
\label{eq:inference}
\end{equation} 

Each instantiation of the expression $Pr(Y = y, e)$ can be computed by summing up all joint entries 
%out all entries in the joint 
that correspond to assignments consistent with $y$ and the evidence variable~$e$. 
The set of random variables $W$ corresponds to variables that are neither query nor evidence.
The~$\alpha$ parameter specifies the normalization factor for distribution~$Pr(Y,e)$, and this normalization factor is informed by certain assumptions made in Bayes rule~\citep{russel10}.

\subsection{Structure Learning in Bayesian Networks}
\label{subsec:learning}

%If the network representation~$\mathcal{G}$ is known, the task of learning the parameters $\Theta$ is a trivial problem, however when $\mathcal{G}$ is known 
%conditional probability tables, which are represented by a set of parameters~$\Theta$. 
%that represent conditional dependencies 
%In this case, given a complete dataset~$\mathcal{D}$, one can simply count how many times each dependency occurs in $\mathcal{D}$ and normalise. For an incomplete dataset, the parameter estimation can be obtained though optimization algorithms such as the \textit{Expectation/Maximization} algorithm~\cite{XXX}. %, which is also a trivial optimisation process. 
%The true problem in learning in Bayesian networks occurs when the network representation $\mathcal{G}$ is unknown and needs to be learned directly from the data together with the respective conditional probability table parameters~$\Theta$.

Bayesian networks are made of two important components: a directed acyclic graph~$\mathcal{G}$ representing the network structure, and a set of probability parameters~$\Theta$ representing the conditional dependence relations. 
Learning a BN is a challenging problem when the network representation $\mathcal{G}$ is unknown. %More specifically, 
Given a dataset $\mathcal{D}$ with $m$ observations, $Pr \left( \mathcal{G}, \Theta | \mathcal{D} \right)$ is composed of two steps, structure learning and parameter learning, as follows~\cite{Scutari19}: \vspace*{-.5\baselineskip}

\begin{equation}
    Pr \left( \mathcal{G}, \Theta | \mathcal{D} \right) = \underbrace{ Pr \left( \mathcal{G} | \mathcal{D} \right) }_\text{structure learning} \cdot  \underbrace{ Pr\left( \Theta| \mathcal{G}, \mathcal{D} \right). }_\text{parameter learning}
\end{equation}

%in which case, Bayesian networks needs to be learned from the input data with the respective conditional probability parameters~$\Theta$.
%Therefore, the problem of learning a Bayesian network structure, $\mathcal{G}$, and the respective parameters, $\Theta$, 

% Scutari and Heckerman papers go here
Structure learning aims to find the directed acyclic graph~$\mathcal{G}$ by maximising $Pr(\mathcal{G} | \mathcal{D})$. Parameter learning, on the other hand, focuses on estimation of the parameters $\Theta$ given the graph $\mathcal{G}$ obtained from structure learning. According to~\cite{Heckerman95,Heckerman95Bayesian}, considering that parameters $\Theta$ represent independent distributions (as assumed in Na\"{i}ve Bayes), the learning process can be formalised as follows~\cite{Scutari19}: \vspace*{-\baselineskip}

%that represents the dependence structure of the data 
%Parameter learning, on the other hand, consists in estimating the parameters $\Theta$ given the graph $\mathcal{G}$, which was obtained from structure learning. Following~\cite{Heckerman95}, if one assumes that parameters represent independent distributions (as it is assumed in Na\"{i}ve Bayes), then the learning process can be formalised in the following way, 

\begin{equation}
Pr( \Theta | \mathcal{G}, \mathcal{D} ) = \prod_i Pr( \Theta_{X_i} | \Pi_{X_i}, \mathcal{D}).
\end{equation}

It is important to note that structure learning is well known to be both NP-hard~\citep{Chickering94} and NP-complete~\citep{Chickering96} due to the following equation: \vspace*{-.5\baselineskip}

\begin{equation}
    Pr( \mathcal{G} | \mathcal{D} )  \propto  Pr( \mathcal{G} ) Pr( \mathcal{D} | \mathcal{G} ),
\end{equation}

\noindent 
which can be decomposed into: \vspace*{-\baselineskip}

\begin{equation}
\begin{split}
    Pr( \mathcal{D} | \mathcal{G} ) = \int Pr( \mathcal{D} | \mathcal{G}, \Theta) Pr(\Theta | \mathcal{G}) d\Theta ~~~~~~~~~~~~~~~~~~~ \\
   ~~~~~~~~~~~~~~~~~~~~~~~~ = \prod_i \int Pr( X_i | \Pi_{X_i},  \Theta_{X_i} ) Pr( \Theta_{X_i} | \Pi_{X_i} ) d\Theta_{X_i}
\end{split}
\end{equation}

In structure learning, it is often used the BIC score, a frequentist measure, to maximise, $Pr( \mathcal{G}, \Theta | \mathcal{D})$, due to its simplicity. \vspace*{-.25\baselineskip}

\begin{equation}
    Score(\mathcal{G}, \mathcal{D}) = BIC( \mathcal{G}, \theta | \mathcal{D} ) = \sum_i log Pr( X_i | \Pi_{X_i}, \Theta_{X_i}) - \frac{log(n)}{2} \left| \Theta_{X_i} \right|.
\end{equation}

% from Scutari 29
According to~\citet{Scutari19}, structure learning via score maximisation is performed using general-purpose optimisation techniques, typically heuristics, adapted to take advantage of these properties to increase the speed of structure learning. The most common are greedy search strategies that employ local moves designed to affect only few local distributions, to that new candidate DAGs can be scored without recomputing the full $Pr(\mathcal{D} | \mathcal{G})$. This can be done either in the space of the DAGs with hill climbing and tabu search~\cite{russel10}. In this paper, we opted for a greedy Hill Climbing approach to learn the structure $\mathcal{G}$, due to its simplicity and effective results~\cite{Heckerman95}.

\subsection{Local Interpretation-Driven Abstract Bayesian Network (LINDA-BN)}
\label{subsec:model}

State-of-the-art techniques for constructing predictive models underpinned by machine intelligence usually adopt a `black-box' approach, where the reasoning behind the predictions remains opaque (particularly in regard to deep learning models). Consequently, the underlying predictive mechanisms remain largely incomprehensible to the decision-maker. The challenge is how to endow machine intelligence with capabilities to explain the underlying predictive mechanisms in a way that helps decision-makers understand and scrutinize the machine learned decisions. In the following, we propose an extended framework of Bayesian Networks for generating post hoc local interpretations of black-box predictive models. We name this framework the \textit{Local Interpretation-Driven Abstract Bayesian Network} (LINDA-BN). It supports extracting a Bayesian network as an approximation (or an abstraction) of a black-box model for a specific prediction learned from any given input. Note that explanations can be constructed from the graphical representations of LINDA-BN, and we will address the explanation generation component as a direction for future work.

% The reason why we refer to the extracted model as \textit{abstract} is simply due to the fact that we are computing a model which is an \textit{approximation} (or an abstraction) of a predictive model represented by the black-box predictor. Although explanations can be easily constructed from these graphical interpretations, we leave the explanation generation component as a future work.

%Given an input vector $\vec{X}$ and a black-box predictive model $\hat{y}$, LINDA-BN mainly consists of three steps: 
%i) to generate a set of permutations $\vec{X}'$, 
%ii) to predict using the permuted points and record the predictions in $\hat{y}(\vec{X}')$, and iii) to learn a Bayesian network that represents the conditional dependencies between the features of $\vec{X}'$ and predictions $\hat{y}(\vec{X}')$. 

\begin{figure}[!h]
    \resizebox{\columnwidth}{!} {
    \includegraphics{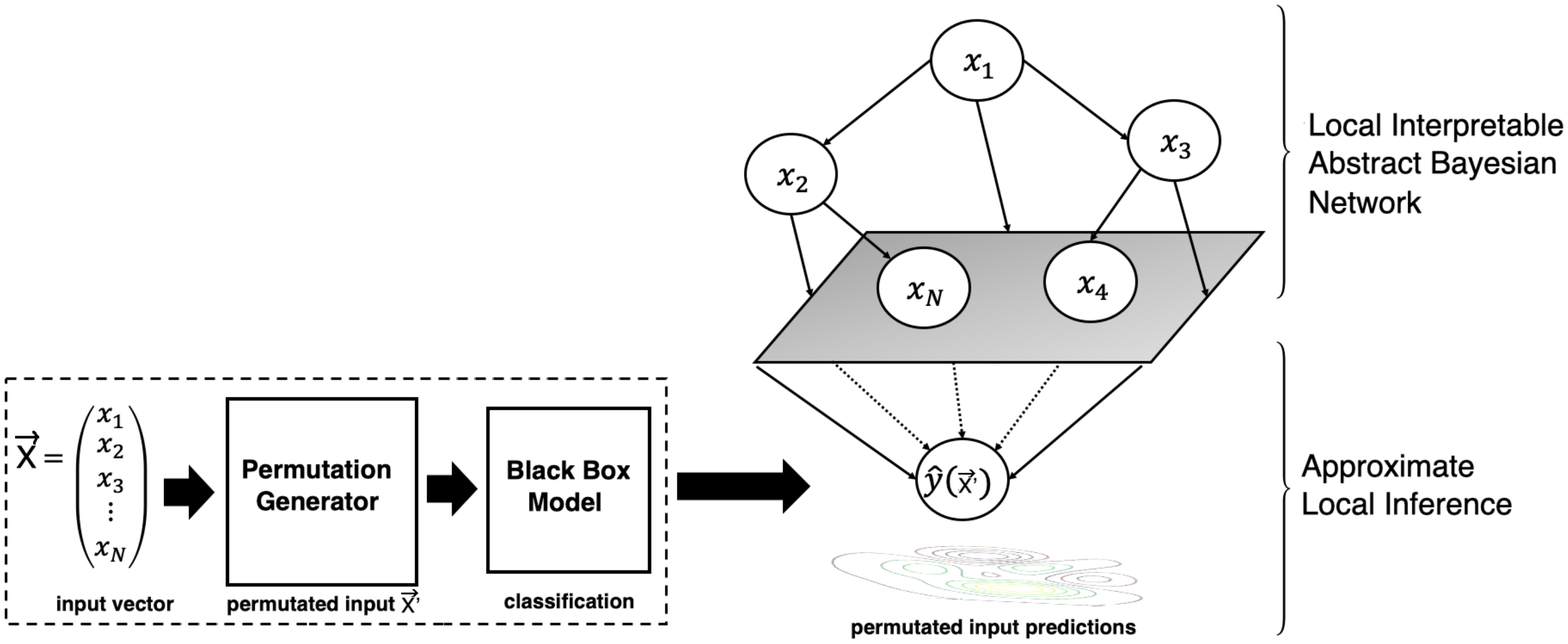}
    }
    \caption{An general illustration of the proposed framework LINDA-BN}
    \label{fig:lia-bn}
\end{figure}

The basic idea behind the proposed framework LINDA-BN rests in three main steps: 
i) permutation generation, ii) Bayesian network learning, and iii) computation of the Markov Blanket of the class variable (representing result of a prediction). It is important to stress that the proposed model aims to augment a decision-maker's intelligence towards a specific decision problem, providing interpretations that can either reinforce the predictions of the black-box or lead to a complete distrust in these predictions (identification of misclassifications). Figure~\ref{fig:lia-bn} shows a general illustration of the proposed framework. % model.

Given a vector of input features $\vec{X} = \{ x_1, x_2, ..., x_n \}$ and a black-box predictor, $\hat{y}(\vec{X})$, the goal is to introduce a set of permutations $\vec{X_i}'$ in the features of $\vec{X}$ in a permutation variance $\epsilon \in \left[ 0, 1 \right]$ in such a way that each feature will be permuted using a uniform distribution over the interval $\left[ x_i - \epsilon, x_i + \epsilon  \right]$. The goal is to analyse how introducing a small perturbation can impact the predictions of the black box prediction, $\hat{y}(\vec{X_i}')$, generating a new statistical distribution describing small variations of the input vector $\vec{X}$. The goal is to learn a Bayesian network structure out of this statistical sample using a Greedy Hill Climbing approach. Our hypothesis is the following: if the data point falls within the correct decision region of the black box predictor, leading to a correct class classification, $c$, then the predictions of all the permutations, $\hat{y}(\vec{X_i}')$, should be close to certainty, i.e. favouring one of the assignments of the class variable with $Pr( Class = c~|~X_i') \approx 1$. This can strengthen the decision-maker's trust in the predictions of the black-box predictor. If, however, the data point, $\vec{X}$, is very close to the black-box's decision boundary, then one would expect that the permutations will be spread around the different regions demarcated by the decision boundary, leading to a more diversified statistical distributions of predictions, and a higher uncertainty in the classification of the respective class $Pr( Class = c ~|~ X_i') << 1$. Such situations have the potential to alert the decision-maker that the black-box predictor is not very certain about the classification of the given data point. Section~\ref{subsec:rules} is centered in this topic.

\begin{algorithm} [!h]
\caption{Local Interpretation-Driven Abstract Bayesian Network Generator}
\label{alg:algorithm}
\begin{algorithmic}[1]
\REQUIRE $local\_vec$, single vector from which we want to generate interpretations  \\
~~~~~~~~~~$black\_box$, a predictive model \\
~~~~~~~~~~$\epsilon$, variance range to permute the features (default = 0.1) \\
~~~~~~~~~~$n\_samples$, number of permuted samples to generate (default = 300) \\
~~~~~~~~~~$class\_var$, string with the name of the class variable \\
	
\ENSURE $\mathcal{G}$, the Local Interpretable Abstract Bayesian Network\\
~~\\
\STATE /* Generate permutations via a uniform distribution within a permutation range  */
\STATE perms = \textit{GeneratePermutations}( x, model, $\epsilon$, n\_samples ) \\
\STATE ~\\
\STATE /* Discretise continuous features according to the number of quartiles  */
\STATE perms\_discr = \textit{DiscretisePermutations}( perms, quartiles = 4 )
\STATE ~\\
\STATE /* Learn BN from discrete permutations using a Greedy Hill Climbing Search */
\STATE bn = \textit{LearnBN\_GreedyHillClimbing}( perms\_discr )
\STATE ~\\
\STATE /* Compute BN's marginal distributions */
\STATE bn\_inf = \textit{ComputeMarginalDistributions}( bn )
\STATE ~\\
\STATE /* Compute BN's Markov blanket */
\STATE bn\_markov = \textit{ComputeMarkovBlanket}( bn, class\_var )
\STATE ~~
\IF{ $bn.nodes <= 10$  } 
\RETURN  $bn\_inf$ /* return full network */ \\
\ELSE 
\RETURN $bn\_markov $ /* return Markov blanket */
\ENDIF
\STATE~~\\ 
\end{algorithmic}
\end{algorithm}

Since the network structure shows dependencies between the input features and the class variable, then it is possible to extract what features contributed to the prediction and \textit{why}, allowing a deeper understanding about the impact that the features have in the class variable, or even provide the decision-maker additional insights about the decision problem. 

For complex decision problems with a large amount of features, the local interpretable network that is learned from the generated permutations is extremely complicated to be analysed by a human, so a Markov Blanket is returned, instead, as a summarisation of what are the main variables influencing the class variable. The Markov blanket determines the boundaries of a system in a statistical sense. It includes all its parents, children, and the parents of the children. 

It can be shown that a node is conditionally independent of all other nodes given values for the nodes in its Markov blanket. Hence, if a node is absent from the class attribute's Markov blanket, its value is completely irrelevant to the classification~\citep{koller09prob}. Algorithm~\ref{alg:algorithm} describes the algorithm that we used for to generate the proposed local interpretable abstract Bayesian network.

\subsection{Interpreting Graphical Representations through Reasoning}
\label{subsec:repr}

This section analyses how to interpret the different situations where a random variable can influence another in the local interpretable Bayesian network model. 

\begin{figure}[h!]
    \resizebox{\columnwidth}{!} {
    \includegraphics[scale=0.1]{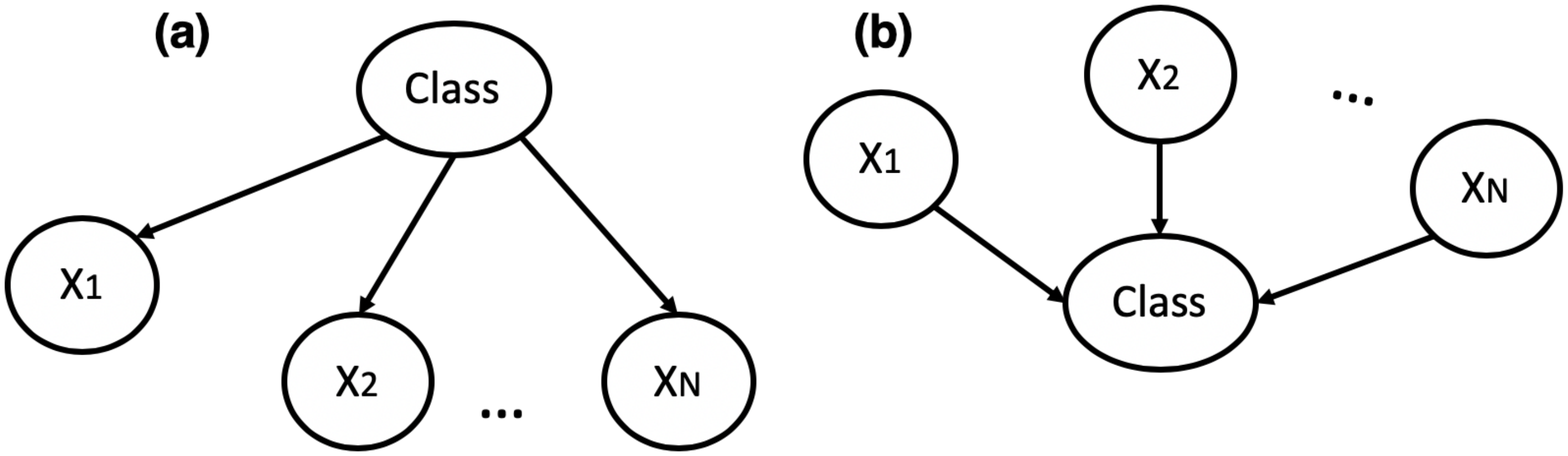}
    }
    \caption{Different graph structures for probabilistic reasoning}
    \label{fig:reasoning}
\end{figure}

In a common cause structure, Figure~\ref{fig:reasoning} (a), the local interpretable model approximates to a Na\"{i}ve Bayes classifier, which means that having knowledge about the class variable will make the feature variables $X_1, X_2, \cdots, X_N$ conditionally independent, and consequently uncorrelated. This means that knowing about $X_1$ does not bring any additional information to the decision-maker. Although human decision-makers tend to assess and interpret these structures as cause/effect relationships as a way to simplify and linearise the decision problem due to bounded rationality constraints, statistically, common cause structures do not imply causal effects in Bayesian networks~\citep{Pearl88}. The consideration of the class variable as a prior in interpretations for a \textit{single datapoint} may indicate a high uncertainty obtained in the statistical sample of the permuted features, suggesting that the datapoint that is being interpreted may be very close to the predictive black-box decision boundary (Section~\ref{subsec:rules} addresses this with a higher detail).

The other type of structure that one can often find in the local interpretable model is the $v$-structure, also called common effect (Figure~\ref{fig:reasoning} (b), which approximates to a linear regression representation. This means that, the features become conditionally independent of the class, if and only if one has knowledge about the class variable. Being uncertain about the class will lead to an influence from the features, $X_1, X_2, \cdots, X_N$. In terms of the proposed local interpretable model, this means that the features have a direct effect in the class variable, and humans can interpret it through an abductive reasoning process.

Abduction is a mode of human reasoning, which was brought into prominence by the the American philosopher C.S. Peirce \citep{Gabbay:Woods:2006}.
In Peirce's view abduction is an inference of the form: ``The surprising fact $C$ is observed. But is $A$ were true, $C$ would be a matter of course. Hence, there is reason to suspect that $A$ is true".
Abductive inference is thus a process of justifying an assumption, hypothesis or conjecture in producing the class of interest.
Peirce states that abduction might explain a given set of data, or might facilitate observationally valid predictions, or might permit the discounting of other hypotheses. 
By engaging in abduction, the decision maker interpreting the graph structure is afforded a \emph{simpler} and \emph{more compact} account \cite{Gabbay:Woods:2006}.

Abduction is not a sound form of inference like deduction, and so even though the decision maker might suspect $A$, there is a degree of uncertainty.
Abduction is sometimes termed ``inference to the best explanation" where there is no guaranteed certainty in the explanation. 
In other words, given a set of observations, the decision maker uses abduction to find the simplest, most likely and compact explanation from the graph structure.  The Markov blanket of the class variable is a way of supporting the decision maker's abductive reasoning process. 
 
\subsection{Rules for Local Interpretations}  
\label{subsec:rules}

The graphical nature of the proposed framework LINDA-BN enables the identification of certain parts that can help the decision-maker assess the reliability of the predictions of the black box for single datapoints. To this end, we propose a set of four rules that correspond to four different patterns that the proposed model can identify, depending on how close to the decision boundary a datapoint is. By analysing the confidence of the interpretable model with regards to the class variable together with the structure of the network, one can provide useful guidelines to the decision-maker that can be later be used to generate human centric and understandable explanations (which is not the focus of this work).

The proposed rules to assess the confidence of the black box predictions using the proposed framework are the following:
\begin{itemize}

\item \textbf{Rule 1: High confidence in predictions.} \textit{If the black box predicts a class $c$ for a given datapoint, $\vec{X}$, and the class variable is contained in a common-effect structure in $\mathcal{G}$ with a probability $Pr(Class = c) \approx 1$, then the interpretable model, $\mathcal{G}$, supports the prediction of $\vec{X}$ and its respective Markov blanket determines the most relevant features.}

As mentioned in Section~\ref{subsec:repr}, common-effect structures in $\mathcal{G}$ approximate to a linear regression representation in which there is a direct influence from the features to the class. When $Pr(Class = c) \approx 1$, then this means that the datapoint falls in a well defined decision region, as illustrated in Figure~\ref{fig:rule1}. Since the likelihood of the class is close to certainty, the decision-maker can make use of the class' respective Markov blanket for explanation and perform an abductive reasoning process in which the decision-maker will seek to find the simplest and most likely conclusion out of the Markov blanket.

\begin{figure}[!h]
    \centering
    \includegraphics[scale=0.22]{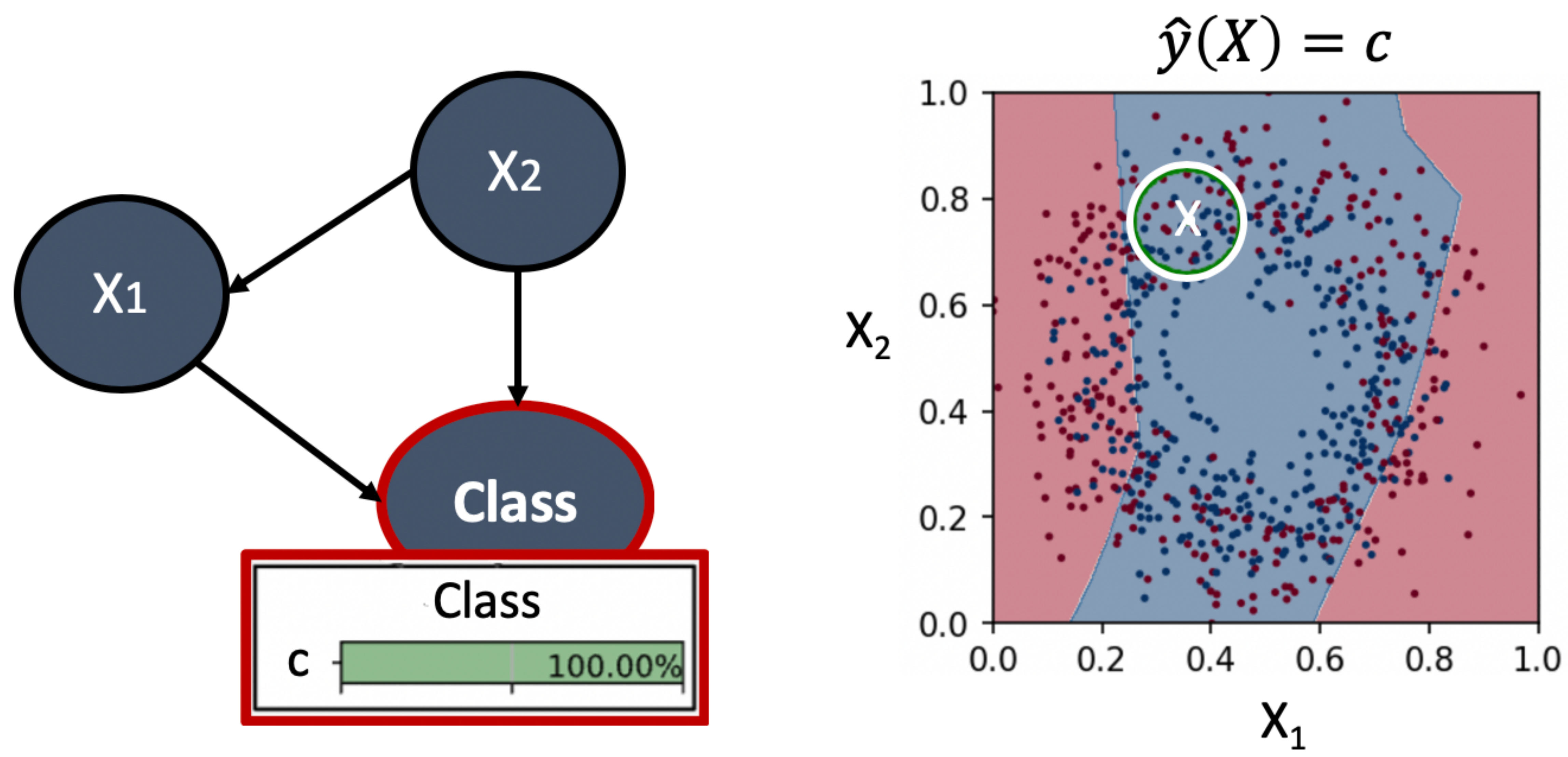}
    \caption{Graphical representation of a pattern representing Rule 1, a high confidence in the prediction of the black box, supported by an interpretable graph showing what are the most relevant features influencing the class variable.}
    \label{fig:rule1}
\end{figure}

\item \textbf{Rule 2: Unreliable predictions.} \textit{If the interpretable network, $\mathcal{G}$, has a structure where the class variable is independent from all other feature variables, that is $ Class \perp \{ X_1, \cdots, X_N \}$, then this corresponds to an unrealistic decision scenario, because the features are uncorrelated from the class variable and providing information about them does not make any change in the probability $Pr( Class = c )$. Thus, the classification $\hat{y}(\vec{X})$ is incorrect and it should be communicated to the decision-maker as an unreliable prediction}.\\

\begin{figure}[!h]
    \centering
    \includegraphics[scale=0.22]{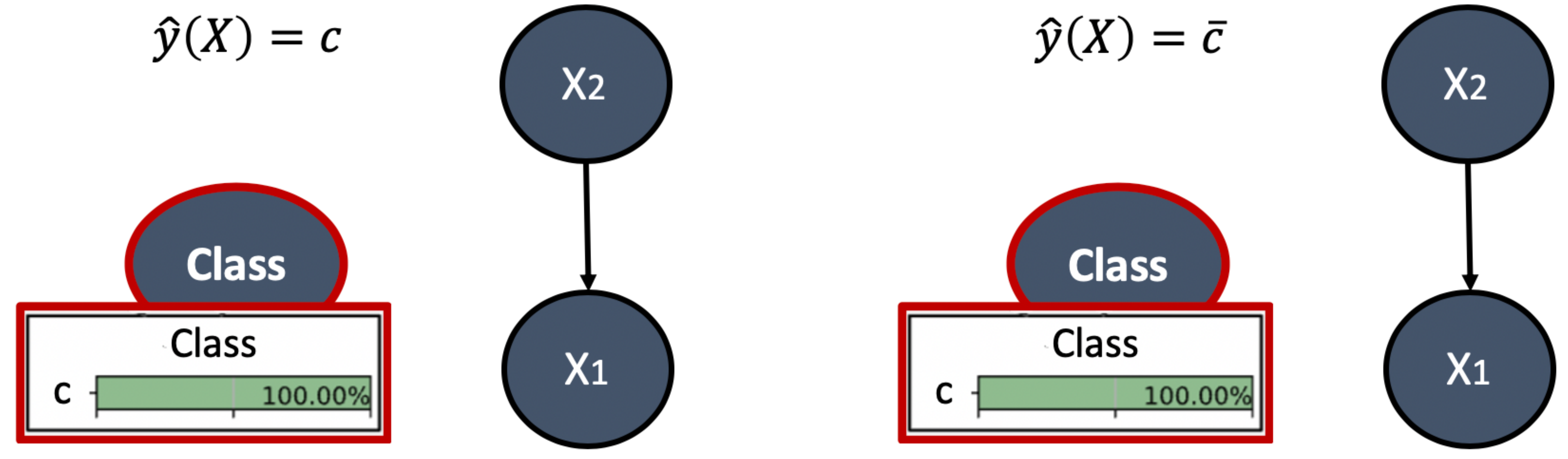}
    \caption{Graphical representation of a pattern representing Rule 2, a distrusted prediction of the black box, supported by an interpretable graph showing that knowing information about the features does not make any changes in the class variable.}
    \label{fig:rule2}
\end{figure}

Sometimes, due to problems in generalising the black box predictor, there can be classifications that are erroneous and unrealistic. In these rare scenarios, the Local Interpretable model can learn from the permuted instances of $\vec{X}$, a graphical structure in which $ Class \perp \{ X_1, \cdots, X_N \}$ (Figure~\ref{fig:rule2} shows an example). In these situations, the Markov Blanket contains only the class variable, which makes it easy to identify the independence in the class variable. Moreover, it can be easily concluded that the classification $\hat{y}(\vec{X})$ is incorrect and it should be communicated to the decision-maker as an unreliable and unrealistic prediction that results from poor generalisation of the black box. % NOTE to match the figure.

\item \textbf{Rule 3: Contrast Effects.} \textit{If the black box predicts $\hat{y}(\vec{X}) = c$, and the maximum likelihood of the class variable in $\mathcal{G}$ is $Pr( Class = \bar{c})$, then there is a contradiction between the local interpretable abstract model and the prediction computed by the black box, suggesting that the datapoint is very close to the decision boundary, which can either be correctly or incorrectly classified. Thus, the decision-maker should analyse the Markov Blanket of the class variable representing $\vec{X}$, and assess whether the relationships between the features justify the class.} \\

\begin{figure}[!h]
    \centering
    \includegraphics[scale=0.22]{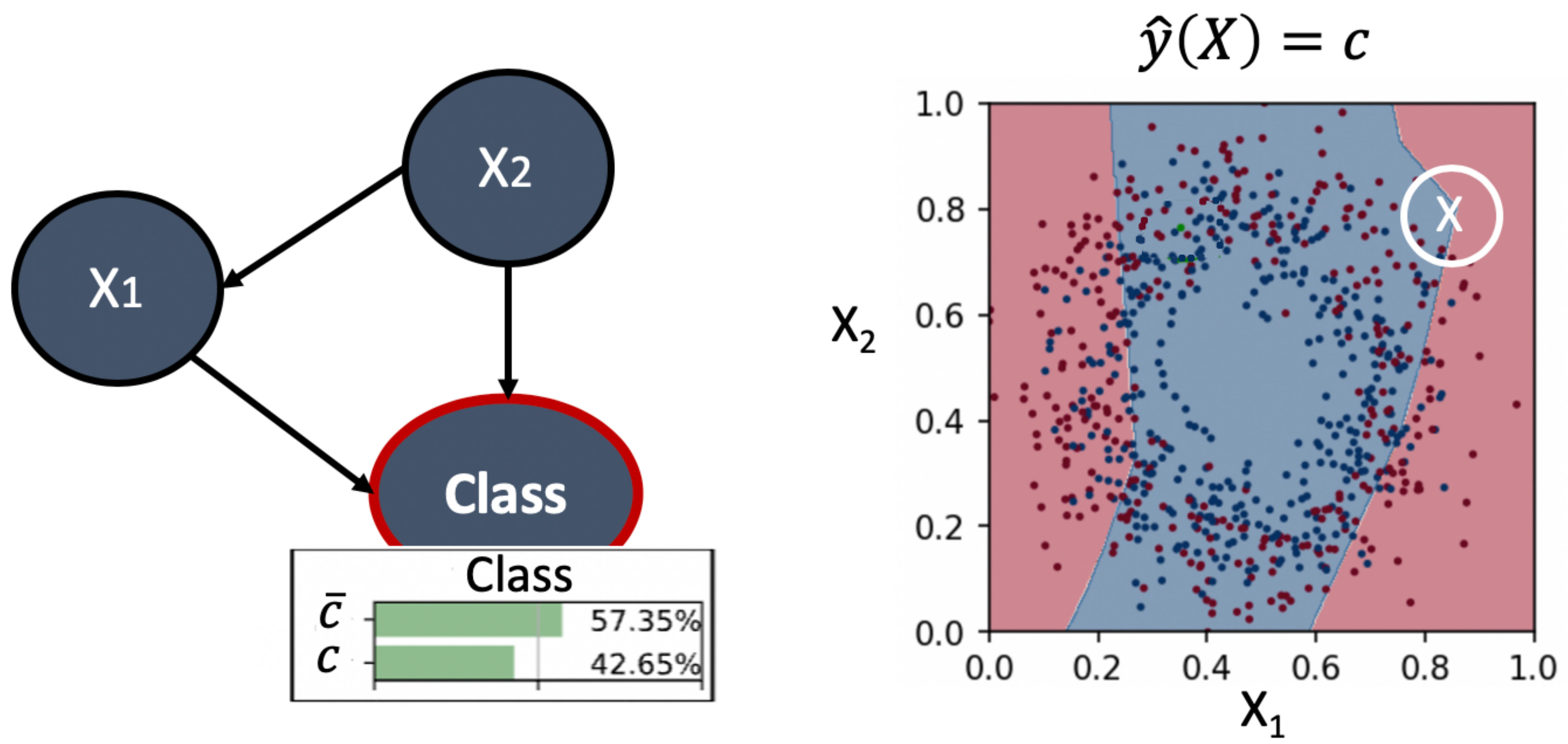}
    \caption{Graphical representation of a pattern representing Rule 3, a contrast effect, where the local interpretable abstract model reinforces a class that is different from the one predicted by the black-box.}
    \label{fig:rule3}
\end{figure}

In situations where the datapoint is very close to a decision boundary, the permutation of the datapoint $\vec{X}$ will generate a statistical distribution within a certain neighbourhood of $X$. Due to the complexity and non-linearity of the decision boundary, the statistical distribution can increase the likelihood, $Pr( Class = \bar{c} )$, contradicting the prediction of the black box, $\hat{y}(\vec{X}) = c$. In these situations, even if the black-box managed to predict correctly $X$, the is a high uncertainty in the prediction, and it should be recommended to the decision-maker to assess the features of $X$ in order to assess its reliability. Figure~\ref{fig:rule3} shows an example of a contrast effect.

\item \textbf{Rule 4: Uncertainty in predictions.} \textit{If the black box predicts $\hat{y}(\vec{X}) = c$, and the probability of the class variable in $\mathcal{G}$ is $Pr( Class = c ) << 1$, but still with a maximum likelihood favouring class $c$, then datapoint $X$ falls near the decision boundary. Even if the class is in accordance with the prediction of the black-box, then there is an underlying uncertainty attached to its prediction. Thus, the decision-maker should analyse the Markov Blanket of the class variable representing $\vec{X}$, and assess whether the relationships between the features justify the class.}

\begin{figure}[!h]
    \centering
    \includegraphics[scale=0.22]{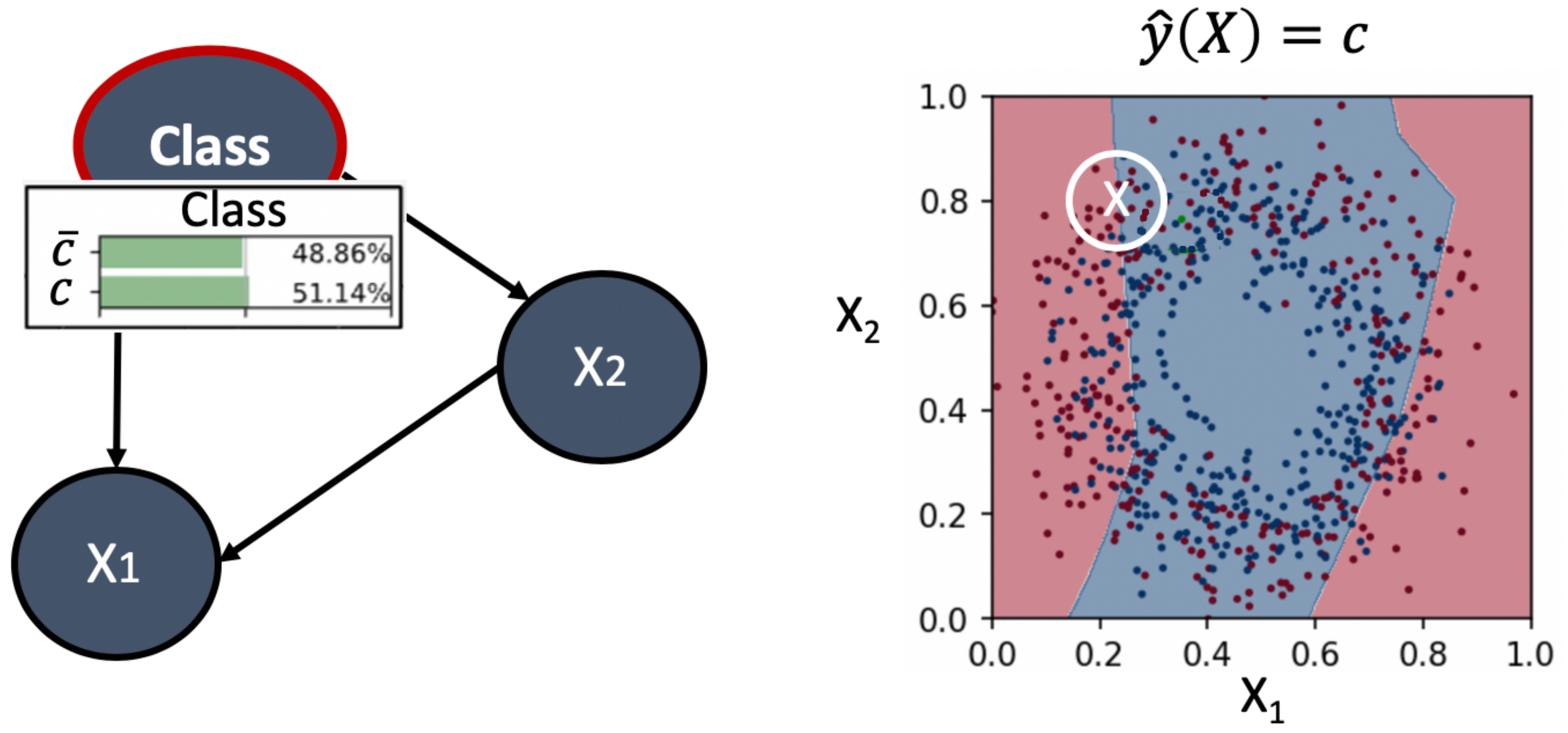}
    \caption{Graphical representation of a pattern representing Rule 4, uncertainty in the prediction, where the local interpretable abstract model shows that the black box prediction is as good as flipping a coin.}
    \label{fig:rule4and5}
\end{figure}

This situation is very similar to the contrast effect (rule 3) with the difference that the class variable in $\mathcal{G}$ is still consistent with the predictions of $\hat{y}(\vec{X})$. However, the statistical distribution of the predictions of the permutations of $\vec{X}$ have a high uncertainty and do not allow the decision-maker to be fully confident in the prediction of $\hat{\vec{X}}$. Thus, depending on the degree of uncertainty of $Pr(Class = c)$, the decision-maker should analyse the Markov Blanket of the class variable representing $\vec{X}$, and assess whether the relationships between the features justify the class. Figure~\ref{fig:rule4and5} shows an example of an uncertain prediction. Although the likelihood of the variable $Class$ is in accordance with $\hat{\vec{X}} = c$, the local interpretable abstract model shows full uncertainty in the prediction: the prediction is as good as flipping a coin.

\end{itemize}

\section{Evaluation}  \label{sec:eval}

Given that there are no standard evaluation metrics for XAI~\citep{guidotti2018}, in this , we present a thorough analysis of the proposed LINDA-BN model in accordance to the rules that we put forward in Section~\ref{subsec:rules}. We performed an analysis in terms of two public well-known datasets from the literature, namely the \textit{Pima Indians diabetes dataset} and the Breast Cancer Wisconsin~\citep{Piri18}, both from the  \textit{UCI Machine Learning Repository}\footnote{\url{https://archive.ics.uci.edu/ml/index.php}}. We have made available a public repository with Jupyter notebooks with the proposed model and all the experiments that we made for this research work: \url{https://github.com/catarina-moreira/LINDA_DSS}.

In Section~\ref{sec:setup}, we present the main experimental setup for our analysis. Section~\ref{sec:params}, presents an analysis of the impact of the permutation variance in the proposed LINDA model. In Section~\ref{sec:analysis}, we make a statistical analysis of the distribution of the interpretations generated by LINDA over both datasets and their the different rules together with existing interpretable approaches such as LIME~\citep{Ribeiro16} and SHAP~\citep{Lundberg17}. Finally, Section~\ref{sec:complex}, describes how the proposed interpretable model performs in more complex decision scenarios.

\subsection{Design of Experiments}\label{sec:setup}

In order to assess the performance and interpretations generated by the proposed LINDA model, we trained a deep learning neural network for two two public well-known datasets from the literature, namely the \textit{Pima Indians diabetes dataset} and the\textit{ Breast Cancer Wisconsin} datasets. Both datasets are highly unbalanced, and for that reason, we had to balance the datasets in order to not have a biased predictive model. 

We performed a grid search approach in order to find the best performing neural network model. The characteristics of the models can be found in Table~\ref{tab:models}. As such, the learned models apply sophisticated internal working mechanisms and run as a black-box when making predictions.

\begin{table}[!h]
    \centering
    \begin{tabular}{l|c|c }
        \textbf{Parameters}             & \textbf{Diabetes}    & \textbf{Breast Cancer}  \\
        \hline
         Model Accuracy                 & 0.7380      & 0.9840 \\
         \hline
         Num. Hidden Layers             & 5           & 4 \\
         \hline
         Num. Neurons per Hidden Layer  & 12          & 12 \\
         \hline
         Hidden Layer Activation function & Relu      & Relu   \\
         \hline
         Ouput Layer Activation function  & Softmax   & Softmax  \\
         \hline
    \end{tabular}
    \caption{Deep neural network architecture found for the best performing model in the Diabetes and the Breast Cancer datasets.}
    \label{tab:models}
\end{table}

\subsection{Analysis of the Impact of Different Permutation Variances}\label{sec:params}

\begin{figure}[!h]
    \resizebox{\columnwidth}{!} {
    \includegraphics{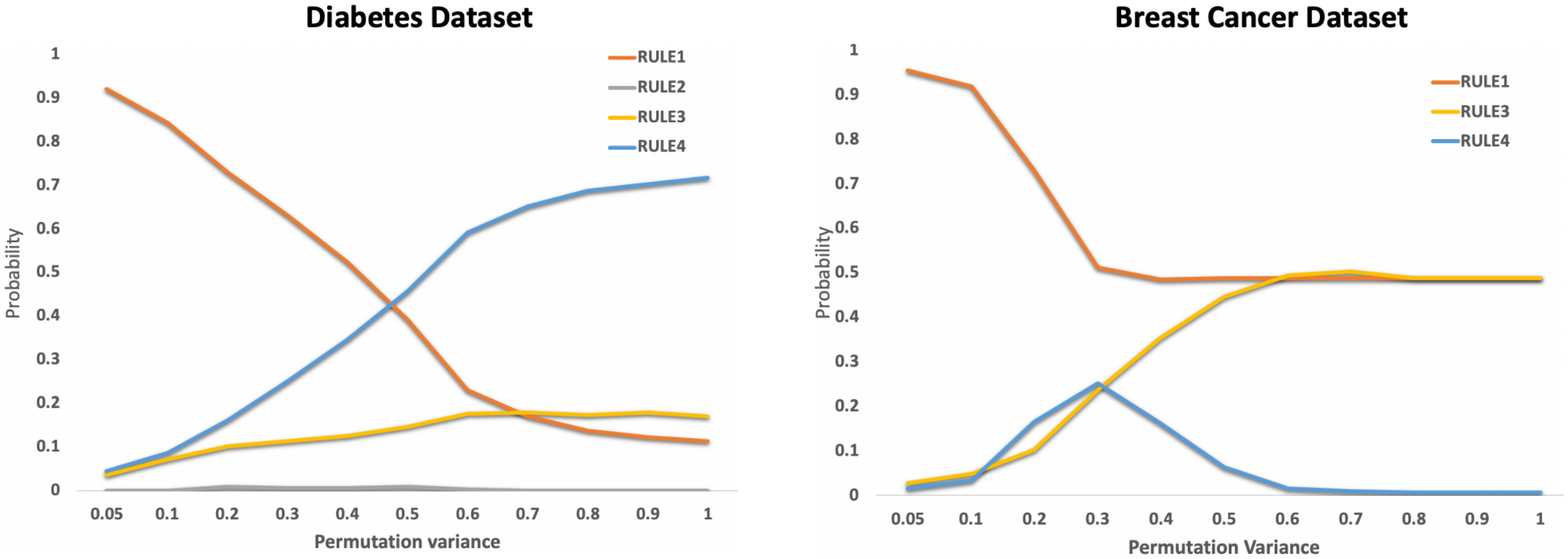}
    }
    \caption{Impact of the permutation variance boundary in the different proposed rules. }
    \label{fig:permutations}
\end{figure}

As in other representative interpretable models in the literature (like LIME and SHAP), LINDA-BN performs permutations between an interval in the range $\left[x_i - \epsilon, x_i + \epsilon \right]$ on the input vector's features $\vec{X} = \{ x_1, x_2, \cdots, x_N\}$ in order to generate a statistical distribution of how the predictions of the black-box change with the features. To investigate the impact of the permutation variance, $\epsilon$, we performed a set of experiments for both datasets, where we varied $\epsilon \in \left[ 0, 1 \right]$, and analysed how many times the proposed interpretable model returned a structure that is consistent with the rules proposed in Section~\ref{subsec:rules}. The obtained results are summarised in Figure~\ref{fig:permutations}.

Taking a close look at the Diabetes dataset in Figure~\ref{fig:permutations}, results show that low variance makes the interpretable model very confident in its interpretations, with $92\%$ of the datapoints (both from the training set and test set) falling in rule 1 (high confidence in the prediction). This is due to the fact that, for a very small permutation interval, the probability of hitting a decision boundary is very small. This is confirmed with the verification of the low amount of datapoints falling in rule 4 (uncertainty in the predictions) or rule 3 (contrast effects). When the permutation variance starts to increase, the model becomes less certain about the predictions. Consequently, rule 1 starts to decrease exponentially, while rule 4 starts to show a lot in uncertainty in the interpretations generated for the different datapoints. One can see that when the permutation variance reaches half of the feature space (note that we assume that the features of the black-box are scaled between 0 and 1 as in standard machine learning applications), then there comes a point where the uncertainty is so high that the interpretable model stops having confidence in almost 90\% of its interpretations. Additionally, almost $80\%$ of the datapoints start falling in rule 4.

In terms of the impact of the permutation variance for the breast cancer dataset, the scenario tends to be slightly different. Just like in the diabetes dataset, when the permutation variance interval is very small, the statistical distribution of the predictions learned by the proposed LINDA model majorly falls in rule 1. However, when the permutation variance starts to increase together with the uncertainty levels, we start to notice an increase of contrast effects (rule 3), rather than uncertainty in the predictions (rule 4). The reason for this is due to the effectiveness of the black-box. Note that the accuracy of the deep neural network model for the diabetes dataset was $73.80 \%$, while the learned model for the breast cancer dataset achieved an accuracy of $98.40\%$. Since the majority of the datapoints fall in well defined decision regions, when the permutation variance increases, the statistical distribution of predictions will tend to show misclassifications (a statistical distribution more concentrated in the opposite region of the decision boundary). Figure~\ref{fig:permutations} also shows that permutation variances superior to 0.2 do not decrease the certainty of the interpretability of correctly predicted datapoints, however it does increase the number of contrast effects (rule 3), reinforcing again the idea that bigger permutation intervals will make the distributions point towards opposite directions of the decision boundary.

\begin{table}[!h]
   \resizebox{\columnwidth}{!} {
\begin{tabular}{l|c||c|c|c|c||c|c|c|c}
\multicolumn{2}{c}{\textbf{}}   & \multicolumn{4}{|c|}{\textbf{DIABETES}}  
                                & \multicolumn{4}{|c}{\textbf{BREAST CANCER}}    \\
 \hline                           
\multicolumn{1}{c|}{\textbf{Rules}}                   & \textbf{Set}   & \multicolumn{1}{c|}{\textbf{TP}} & \multicolumn{1}{c|}{\textbf{TN}} & \multicolumn{1}{c|}{\textbf{FP}} & \multicolumn{1}{c|}{\textbf{FN}} & \multicolumn{1}{|c|}{\textbf{TP}} & \multicolumn{1}{c|}{\textbf{TN}} & \multicolumn{1}{c|}{\textbf{FP}} & \multicolumn{1}{c|}{\textbf{FN}}  \\
\hline 
\multicolumn{1}{c|}{\multirow{2}{*}{\textbf{Rule 1}}} & \textbf{Train} & 0.8662   & 0.91   & 0.7627   & 0.7419    & 0.9931   & 0.8786   & 1.0000   & 0.1667  \\
\cline{2-10} 
\multicolumn{1}{c|}{}               & \textbf{Test}  & 0.7576   & 0.96    & 0.57   & 0.8571   & 1.0000   & 0.8286    & 1.0000   & 0.0000 \\
\hline
\hline
\multirow{2}{*}{\textbf{Rule 2}}    & \textbf{Train}  & 0.0000    & 0.0000   & 0.0000   & 0.0000   & 0.0000    & 0.0000   & 0.0000   & 0.0000  \\
\cline{2-10} 
                                    & \textbf{Test}  & 0.0000    & 0.0000   & 0.0000   & 0.0000   & 0.0000    & 0.0000   & 0.0000   & 0.0000 \\
\hline
\hline
\multirow{2}{*}{\textbf{Rule 3}}    & \textbf{Train} & 0.0828  & 0.016   & 0.1186   & 0.0645  &0.0000  & 0.0643   & 0.0000   & 0.6667   \\
\cline{2-10} 
                                    & \textbf{Test} & 0.1818  & 0.0000   & 0.14   & 0.0000   & 0.0000  & 0.1143   & 0.0000   & 0.0000  \\
\hline
\hline
\multirow{2}{*}{\textbf{Rule 4}}    & \textbf{Train} & 0.0509  & 0.0787   & 0.1186   & 0.1935  & 0.0069  & 0.0571   & 0.0000   & 0.1667   \\      
\cline{2-10} 
                                    & \textbf{Test} & 0.0606   & 0.04   & 0.29   & 0.1429  & 0.0000   & 0.0571   & 0.0000   & 0.0000    \\ 
\hline
\end{tabular}
}
\caption{Overview of the distribution of the datapoints for the Diabetes and Breast Cancer datasets over the proposed rules using LINDA in order to determine the confidence of the computed interpretations.}
\label{tab:results}
\end{table}

In order to extract interpretable models that are both highly confident in the interpretations, but  can also flag possible misclassifications, we decided to set the permutation variance to $0.1$ for the remaining parts of this analysis.

\subsection{Analysis of Rules for Local Interpretations}\label{sec:analysis}

In this section, we analyse the impact of the proposed rules in the different classifications: true positives, true negatives, false positives, and false negatives. The goal with this analysis is to understand if the proposed rules can provide the decision-maker some insights on whether or not, the decision-maker is facing a correct classification or a possible misclassification. Table~\ref{tab:results}
shows the percentage of datapoints over the different proposed rules for both the diabetes and breast cancer datasets, using $\epsilon = 0.1$.

\begin{itemize}

\item \textbf{Correct classifications majorly coincide with rule 1, leading to highly confident predictions}. For the breast cancer dataset, for instance, all datapoints classified as true positives and true negatives were categorised as rule 1 with high confident interpretations. For the diabetes dataset, since the black box had a more poor performance, then the percentage of correctly classified datapoints is already smaller. Still, 86\% of the true positives in the training set fell under the category of rule 1 and in the test set 75\%. Regarding the true negatives, more than 90\% of the datapoints had interpretations supporting a true classification of a non-diabetes case. When compared with interpretations from state of the art algorithms, one can see that both LIME and SHAP also tend to reinforce the features that are contributing positively to the class diabetes. Figure~\ref{fig:rule1_comp} shows an example of an interpretation of a correctly classified datapoint ($\epsilon = 0.1$) and the respective interpretations for LIME and SHAP. For the case of misclassifications, there was a significant amount of datapoints belonging to the set of the false positives and the false negatives that were also categorised as rule 1. In these examples, the interpretable model could not provide significant insights of why there was a misclassification. 

\begin{figure}[H]
    \resizebox{\columnwidth}{!} {
    \includegraphics{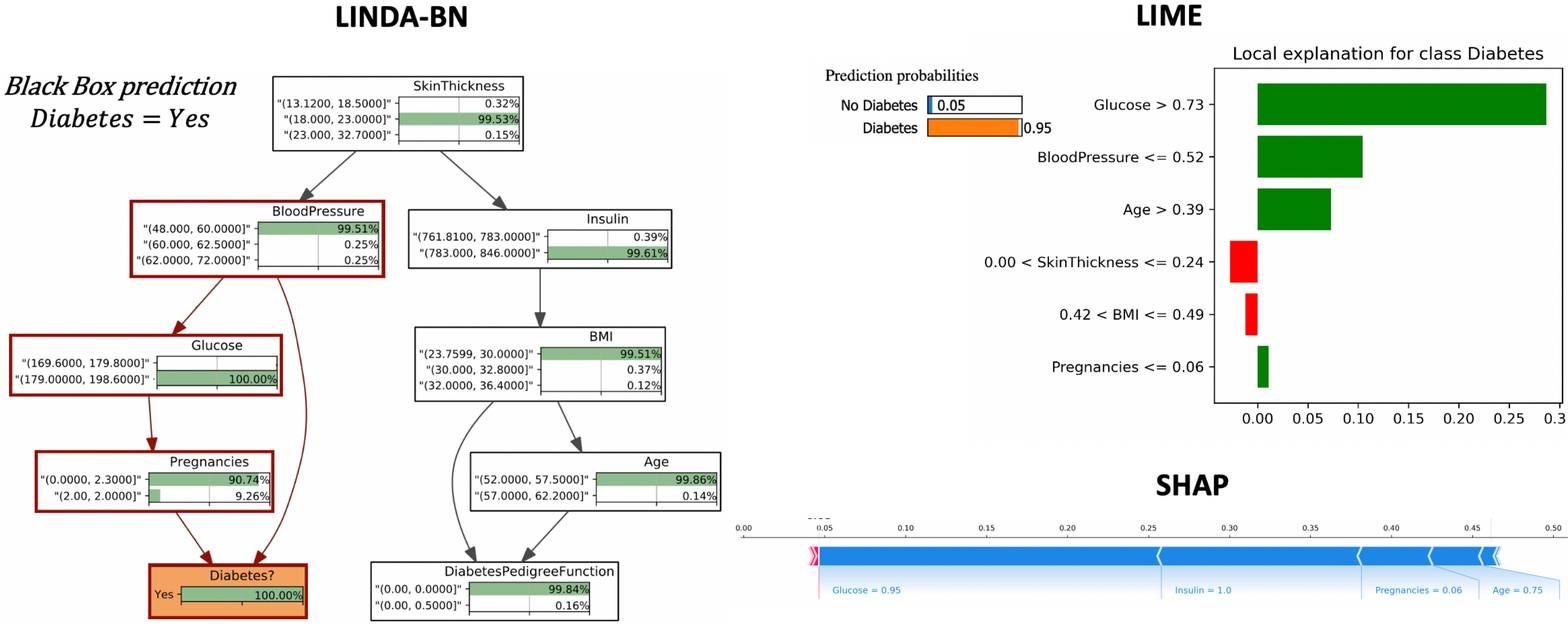}
    }
    \caption{Correct prediction high confidence (a true positive).}
    \label{fig:rule1_comp}
\end{figure}

\item \textbf{Nonexistence of rule 2.} As illustrated in Figure~\ref{fig:permutations}, rule 2, which corresponds to the cases where the class variable is independent of the features, is extremely rare. This rule only starts to emerge for permutations superior to 0.2 in the diabetes dataset, and are nonexistent in the cancer dataset. This suggests that permutation variances of 0.1 do not point towards unrealistic and erroneous classifications. Figure~\ref{fig:rule2_comp} shows an example of an unreliable prediction that was found in the testset of the diabetes dataset, using a variance of $\epsilon = 0.25$ and the respective LIME and SHAP interpretations. Note that this specific datapoint corresponds to a misclassification (a false positive).

\begin{figure}[!h]
    \resizebox{\columnwidth}{!} {
    \includegraphics{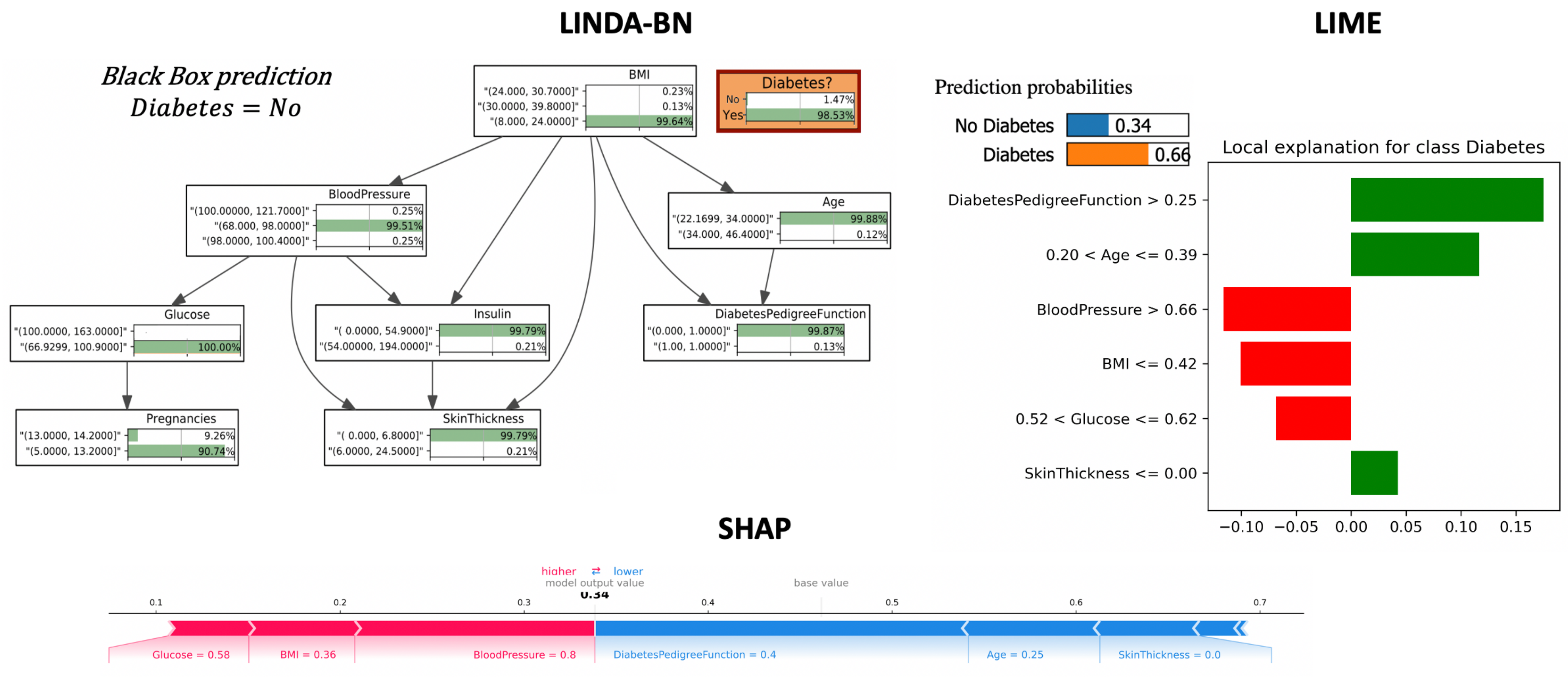}
    }
    \caption{Misclassification in accordance with rule 2, an unreliable prediction (a false positive).}
    \label{fig:rule2_comp}
\end{figure}

\item \textbf{Contrast effects, rule 3, majorly coincide with misclassifications.} When a datapoint falls very close to the decision boundary, then when permuting that datapoint, there can be a significant statistical distribution of the predictions falling on the region of the decision boundary of the opposite class. According to the analysis in Table~\ref{tab:results}, in the Diabetes dataset, rule 3 majorly occurred in the set of false positive points, that is datapoints that were misclassified. But there is also a significant percentage of datapoints in the set of true positives. Although these points were correctly classified, the contrast effect that is captured by the proposed interpretable model might indicate that these are cases correctly classified near the decision boundary. Thus, the decision-maker should be aware that although there was a correct classification, this classification might not have occurred due to the appropriate input feature values. Figure~\ref{fig:rule3_comp} shows an example of a rule 3 in the Diabetes dataset for $\epsilon = 0.1$. When comparing with LIME and SHAP, one can notice that it is not very clear that this datapoint represents a misclassification. 

\begin{figure}[!h]
    \resizebox{\columnwidth}{!} {
    \includegraphics{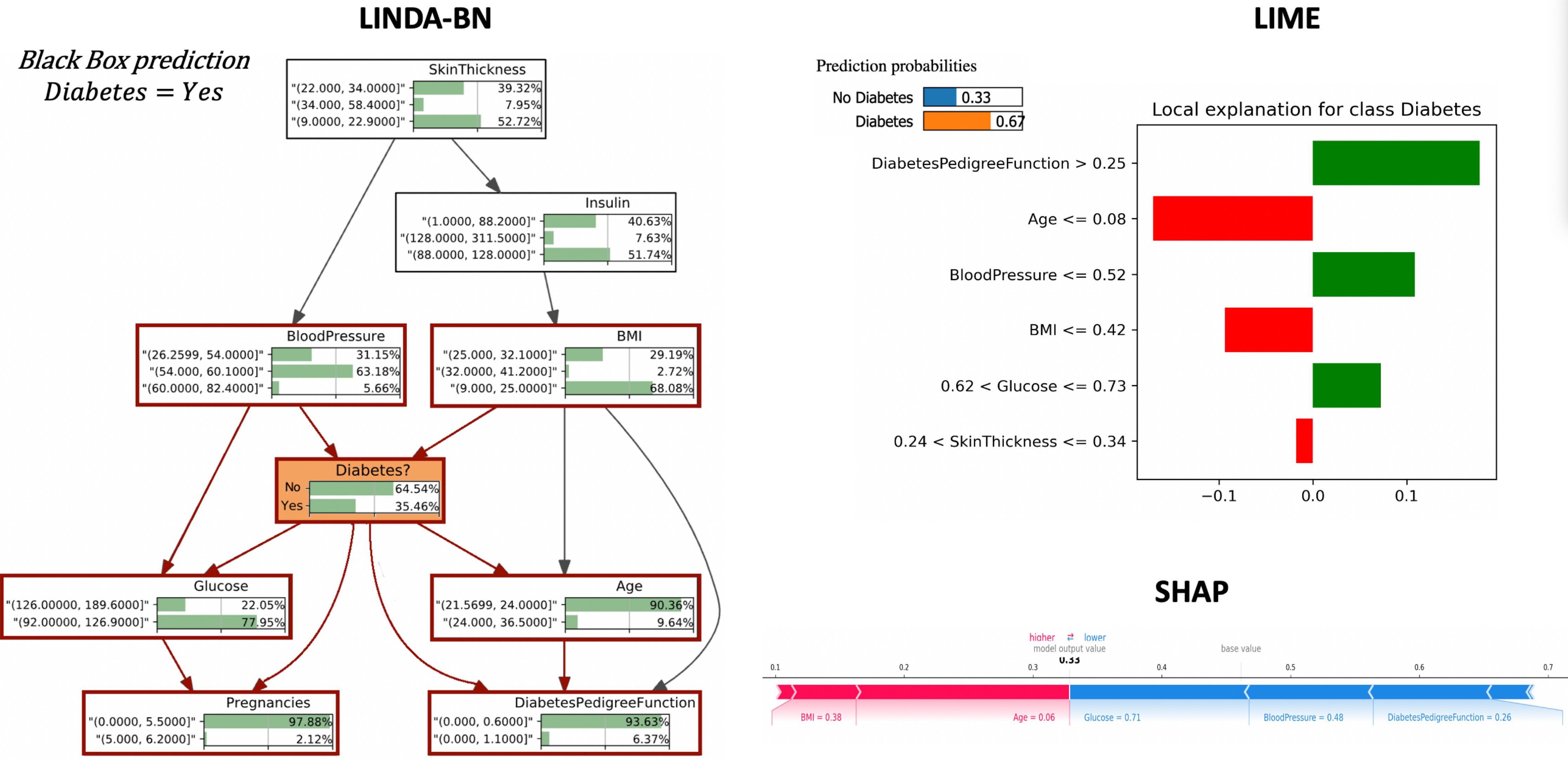}
    }
    \caption{Misclassification in accordance with rule 3, a contrast effect.}
    \label{fig:rule3_comp}
\end{figure}

\item \textbf{Uncertainty in predictions, rule 4, majorly coincide with misclassifications.} In the experiments that were performed, using $\epsilon = 0.1$, Table~\ref{tab:results} shows that the datapoints showing higher uncertainty in the likelihood of the class variable are mostly present in the sets of the false positives and the false negatives. In other words, in the set of misclassified datapoints. This is more noticeable in the diabetes dataset, in which the black-box predictor achieve an average accuracy of $73.8\%$ where there are more misclassified datapoints. On the other hand, when one looks at the breast cancer dataset, since there were almost no misclassified datapoints (accuracy of 0.9840), the percentage of datapoints falling in rule 4 are nearly zero. Figure~\ref{fig:rule4_comp} illustrates an example of a false positive datapoint, in which the interpretable model show maximum uncertainty in the class variable node. In terms of LIME and SHAP, it is hard to identify any principles that could point the decision-maker about a possible misclassification.

\begin{figure}[!h]
    \resizebox{\columnwidth}{!} {
    \includegraphics{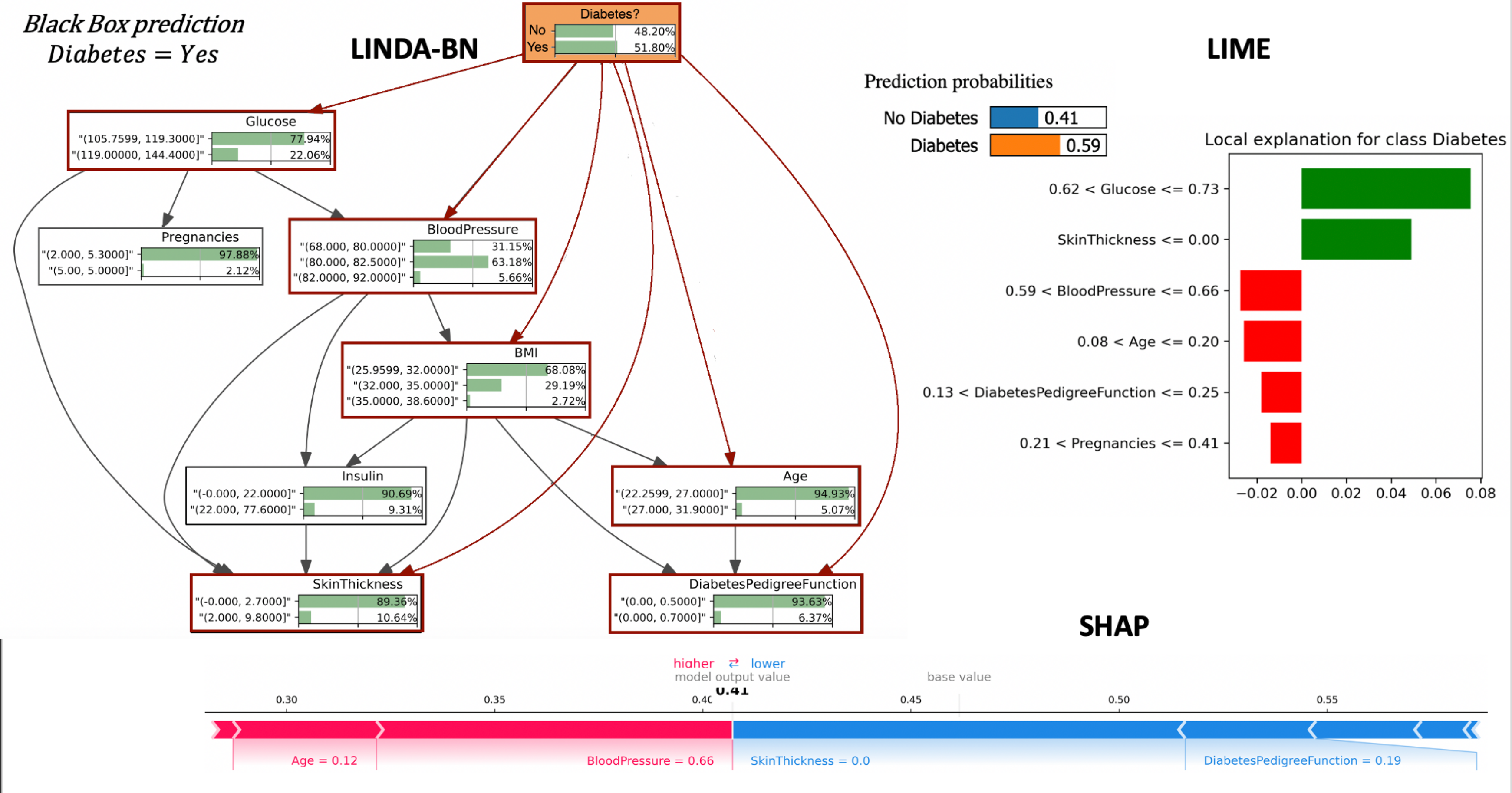}
    }
    \caption{Correct classification in accordance with rule 4, uncertainty in predictions (true positive).}
    \label{fig:rule4_comp}
\end{figure}

\end{itemize}

In the next section, we describe how more complex decision problems are addressed by the proposed interpretable model.

\subsection{Interpretations for Complex Decision Scenarios}\label{sec:complex}

For small decision problems (at most 10 random variables), the proposed LINDA model displays the full interpretable network. For more complex decision problems, however, this would become unreadable for a human decision-maker. The breast cancer dataset is an example of such complex decision problem that contains a set of 30 features, which are mapped into random variables. This results in a graphical structure too complex for any human to analyse. For such datasets, the proposed LINDA model provides the decision-maker a Markov Blanket of the variable of interest, instead of the full local interpretable Bayesian network, together with information about in which rule the network pattern corresponds to and respective marginal probabilities. 
\begin{figure}[H]
    \resizebox{\columnwidth}{!} {
    \includegraphics{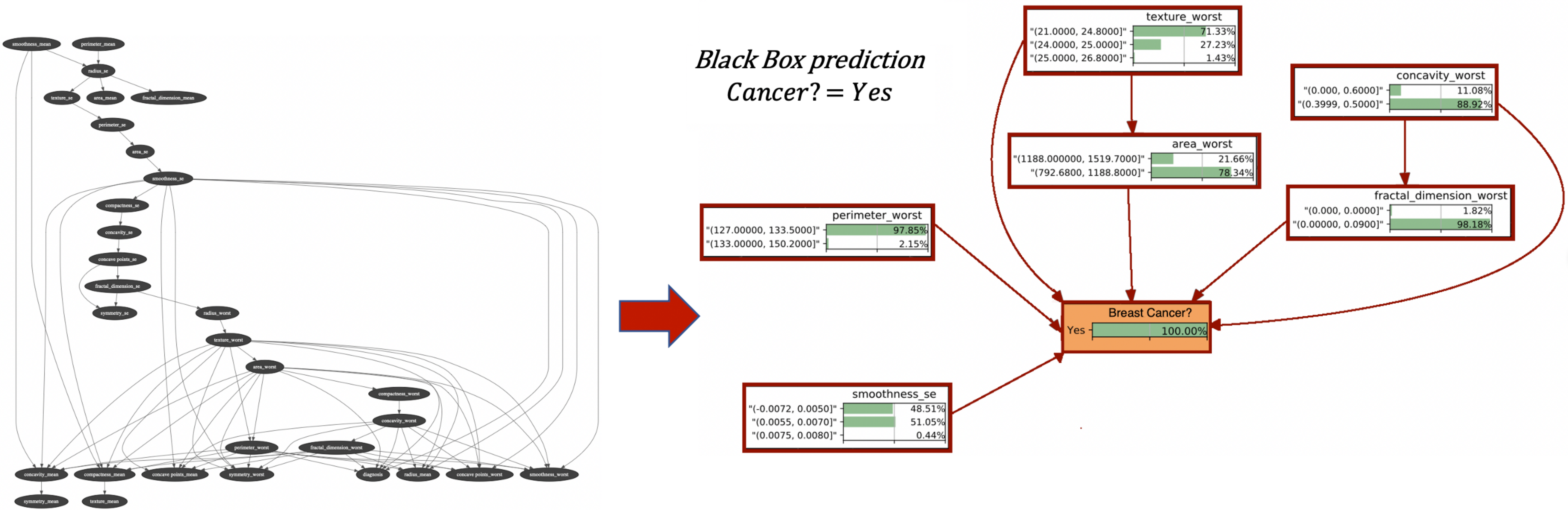}
    }
    \caption{Markov blanket representation of a local interpretable Bayesian network with 30 nodes for the breast cancer dataset. The Markov blanket is in accordance with rule 1 and represents a correct classification.}
    \label{fig:complex}
\end{figure}
This representation enables the summarisation of information, enabling a fast and compact data driven interpretation of a local datapoint. Figure~\ref{fig:complex} shows an example of an interpretable network that was extracted out of a true positive datapoint. The complexity of the network does not enable any human interpretations to take place. However, when looking at the Markov blanket together with the marginal probabilities of the random variables, then one can clearly identify a common-effect structure in which six features directly influence the class variable and contribute to its value. Moreover, the statistical distribution of the permutations shows a full confidence in the diagnosis, suggesting that the datapoint falls within rule 1 and consequently there is a high confidence that it is a correct prediction. The depth of this Markov blanket can potentially be extended to different depths, depending on the decision-maker's needs (for example. a normal person would be satisfied with the Markov blanket in Figure~\ref{fig:complex}, however a medical doctor would probability explore other depths and analyse the relationship between more variables and their indirect influences towards the class variable).

\section{Conclusions}  \label{sec:concl}

In this paper, we proposed a new post hoc interpretable framework, the \textit{Local Interpretation-Driven Abstract Bayesian Network} (LINDA-BN). This framework consists in learning a Bayesian network as an approximation of a black-box model from a statistical distribution of predictions from a local datapoint. 

The major contribution of the proposed framework is the ability to identify four different rules which can inform the decision-maker about the confidence level in a given prediction of a specific datapoint. As such, the interpretations provided in our approach can help the decision-maker assess the reliability of predictions learned by a black-box model. These rules correspond to the different patterns that can be found in the learned Bayesian network, and they are summarised as follows:
\begin{itemize}
    \item Rule 1 - High confidence in predictions: a common-effect structure in which the features are directly influencing the class and the maximum likelihood of the class is close to one, suggesting a correct classification.
    \item  Rule 2 - Unreliable predictions: when the class variable is independent from the features, suggesting a misclassification
    \item  Rule 3 - Contrast Effects: when the maximum likelihood of the class variable in the learned Bayesian network favours a class opposite to the black-box model, suggesting a misclassification.
    \item Rule 4 - Uncertainty in the predictions: when the likelihood of the class variable has very high levels of uncertainty, suggesting that the decision-maker should assess the network in order to understand if the features are supporting the class.
\end{itemize}

Experimental findings showed that rules 3 and 4 usually occurred in sets of false positives and false negatives, suggesting that the proposed framework might provide a possible approach to identify misclassifications in black-box models. On the other hand, the correct classifications (true positives and true negatives), were mostly associated with rule 1 with common-effect graph structures and maximum likelihood in the class variable close to one, again providing a potential method to identify correct classifications and promote trust in the decision-maker.

For future work, we would like to extend the proposed approach from an interpretable framework to an explainable one. This majorly consists in converting the symbolic rules proposed in this study into explainable arguments that could communicate the decision-maker \textit{why} a certain prediction was computed out of a black-box and the reasons of \textit{why / why not} a decision-maker should trust in the predictions.

%\bibliographystyle{model1-num-names} 
%\bibliography{literature}

\begin{thebibliography}{36}
\expandafter\ifx\csname natexlab\endcsname\relax\def\natexlab#1{#1}\fi
\providecommand{\url}[1]{\texttt{#1}}
\providecommand{\href}[2]{#2}
\providecommand{\path}[1]{#1}
\providecommand{\DOIprefix}{doi:}
\providecommand{\ArXivprefix}{arXiv:}
\providecommand{\URLprefix}{URL: }
\providecommand{\Pubmedprefix}{pmid:}
\providecommand{\doi}[1]{\href{http://dx.doi.org/#1}{\path{#1}}}
\providecommand{\Pubmed}[1]{\href{pmid:#1}{\path{#1}}}
\providecommand{\bibinfo}[2]{#2}
\ifx\xfnm\relax \def\xfnm[#1]{\unskip,\space#1}\fi
%Type = Article
\bibitem[{Murdoch et~al.(2019)Murdoch, Singh, Kumbier, Abbasi{-}Asl, and
  Yu}]{corr/abs-1901-04592}
\bibinfo{author}{W.~J. Murdoch}, \bibinfo{author}{C.~Singh},
  \bibinfo{author}{K.~Kumbier}, \bibinfo{author}{R.~Abbasi{-}Asl},
  \bibinfo{author}{B.~Yu},
\newblock \bibinfo{title}{Interpretable machine learning: definitions, methods,
  and applications},
\newblock \bibinfo{journal}{CoRR} \bibinfo{volume}{abs/1901.04592}
  (\bibinfo{year}{2019}).
%Type = Article
\bibitem[{Liao et~al.(2020)Liao, Gruen, and Miller}]{corr/abs-2001-02478}
\bibinfo{author}{Q.~V. Liao}, \bibinfo{author}{D.~M. Gruen},
  \bibinfo{author}{S.~Miller},
\newblock \bibinfo{title}{Questioning the {AI:} informing design practices for
  explainable {AI} user experiences},
\newblock \bibinfo{journal}{CoRR} \bibinfo{volume}{abs/2001.02478}
  (\bibinfo{year}{2020}).
%Type = Article
\bibitem[{Guidotti et~al.(2018)}]{guidotti2018}
\bibinfo{author}{R.~Guidotti}, et~al.,
\newblock \bibinfo{title}{A survey of methods for explaining black box models},
\newblock \bibinfo{journal}{ACM Computing Survey} \bibinfo{volume}{51}
  (\bibinfo{year}{2018}) \bibinfo{pages}{93:1--93:42}.
%Type = Inproceedings
\bibitem[{Lakkaraju et~al.(2019)}]{lakkaraju2019}
\bibinfo{author}{H.~Lakkaraju}, et~al.,
\newblock \bibinfo{title}{Faithful and customizable explanations of black box
  models},
\newblock in: \bibinfo{booktitle}{Proceedings of the 2019 {AAAI} Conference on
  {AIES} 2019}, \bibinfo{year}{2019}, pp. \bibinfo{pages}{131--138}.
%Type = Article
\bibitem[{Lipton(2018)}]{lipton2018}
\bibinfo{author}{Z.~C. Lipton},
\newblock \bibinfo{title}{The mythos of model interpretability},
\newblock \bibinfo{journal}{{CACM}} \bibinfo{volume}{61} (\bibinfo{year}{2018})
  \bibinfo{pages}{36--43}.
%Type = Article
\bibitem[{Doshi-Velez and Kim(2017)}]{doshivelez2017}
\bibinfo{author}{F.~Doshi-Velez}, \bibinfo{author}{B.~Kim},
\newblock \bibinfo{title}{Towards a rigorous science of interpretable machine
  learning},
\newblock \bibinfo{journal}{arxiv: 1702.08608}  (\bibinfo{year}{2017}).
%Type = Article
\bibitem[{Holzinger et~al.(2019)Holzinger, Langs, Denk, Zatloukal, and
  M\"{u}ller}]{Holzinger19}
\bibinfo{author}{A.~Holzinger}, \bibinfo{author}{G.~Langs},
  \bibinfo{author}{H.~Denk}, \bibinfo{author}{K.~Zatloukal},
  \bibinfo{author}{H.~M\"{u}ller},
\newblock \bibinfo{title}{Causability and explainability of artificial
  intelligence in medicine},
\newblock \bibinfo{journal}{Wiley Interdisciplinary Reviews: Data Mining and
  Knowledge Discovery} \bibinfo{volume}{9} (\bibinfo{year}{2019})
  \bibinfo{pages}{e1312}.
%Type = Article
\bibitem[{Siering et~al.(2018)Siering, Deokar, and Janze}]{Siering2018}
\bibinfo{author}{M.~Siering}, \bibinfo{author}{A.~V. Deokar},
  \bibinfo{author}{C.~Janze},
\newblock \bibinfo{title}{Disentangling consumer recommendations: Explaining
  and predicting airline recommendations based on online reviews},
\newblock \bibinfo{journal}{Decision Support Systems} \bibinfo{volume}{107}
  (\bibinfo{year}{2018}) \bibinfo{pages}{52 -- 63}.
%Type = Article
\bibitem[{Kim et~al.(2020)Kim, Park, and Suh}]{KIMDSS2020}
\bibinfo{author}{B.~Kim}, \bibinfo{author}{J.~Park}, \bibinfo{author}{J.~Suh},
\newblock \bibinfo{title}{Transparency and accountability in ai decision
  support: Explaining and visualizing convolutional neural networks for text
  information},
\newblock \bibinfo{journal}{Decision Support Systems} \bibinfo{volume}{134}
  (\bibinfo{year}{2020}) \bibinfo{pages}{113302}.
%Type = Inproceedings
\bibitem[{Ribeiro et~al.(2016)Ribeiro, Singh, and Guestrin}]{Ribeiro16}
\bibinfo{author}{M.~T. Ribeiro}, \bibinfo{author}{S.~Singh},
  \bibinfo{author}{C.~Guestrin},
\newblock \bibinfo{title}{"{W}hy {S}hould {I} {T}rust {Y}ou?": Explaining the
  predictions of any classifier},
\newblock in: \bibinfo{booktitle}{Proceedings of the 22nd ACM SIGKDD
  International Conference on Knowledge Discovery and Data Mining},
  \bibinfo{year}{2016}, pp. \bibinfo{pages}{1135--1144}.
%Type = Inproceedings
\bibitem[{Lundberg and Lee(2017)}]{Lundberg17}
\bibinfo{author}{S.~Lundberg}, \bibinfo{author}{S.-I. Lee},
\newblock \bibinfo{title}{A unified approach to interpreting model
  predictions},
\newblock in: \bibinfo{booktitle}{Proceedings of the 31st Annual Conference on
  Neural Information Processing Systems (NIPS)}, \bibinfo{year}{2017}.
%Type = Inproceedings
\bibitem[{Radwa~Elshawi and Sakr(2019)}]{Elshawi19}
\bibinfo{author}{M.~A.-M. Radwa~Elshawi, Youssef~Sherif},
  \bibinfo{author}{S.~Sakr},
\newblock \bibinfo{title}{Interpretability in healthcare a comparative study of
  local machine learning interpretability techniques},
\newblock in: \bibinfo{booktitle}{Proceedings of IEEE Symposium on
  Computer-Based Medical Systems (CBMS)}, \bibinfo{year}{2019}.
%Type = Inproceedings
\bibitem[{Stiffler et~al.(2018)Stiffler, Hudler, Lee, Braines, Mott, and
  Harborne}]{Stiffler18}
\bibinfo{author}{M.~Stiffler}, \bibinfo{author}{A.~Hudler},
  \bibinfo{author}{E.~Lee}, \bibinfo{author}{D.~Braines},
  \bibinfo{author}{D.~Mott}, \bibinfo{author}{D.~Harborne},
\newblock \bibinfo{title}{An analysis of the reliability of lime with deep
  learning models},
\newblock in: \bibinfo{booktitle}{Proceedings of the Dstributed Analytics and
  Information Science International Technology Alliance}, \bibinfo{year}{2018}.
%Type = Misc
\bibitem[{Tan et~al.(2019)Tan, Song, Udell, Sun, and Zhang}]{Tan17}
\bibinfo{author}{H.~F. Tan}, \bibinfo{author}{K.~Song},
  \bibinfo{author}{M.~Udell}, \bibinfo{author}{Y.~Sun},
  \bibinfo{author}{Y.~Zhang}, \bibinfo{title}{Why should you trust my
  interpretation? understanding uncertainty in lime predictions},
  \bibinfo{year}{2019}.
%Type = Inproceedings
\bibitem[{Ribeiro et~al.(2018)Ribeiro, Singh, and Guestrin}]{Ribeiro18}
\bibinfo{author}{M.~T. Ribeiro}, \bibinfo{author}{S.~Singh},
  \bibinfo{author}{C.~Guestrin},
\newblock \bibinfo{title}{Anchors: High-precision model-agnostic explanations},
\newblock in: \bibinfo{booktitle}{Proceedings of the 32nd AAAI International
  Conference on Artificial Intelligence}, \bibinfo{year}{2018}.
%Type = Article
\bibitem[{Shapley(1952)}]{Shapley52}
\bibinfo{author}{L.~S. Shapley},
\newblock \bibinfo{title}{A value for n-person games},
\newblock \bibinfo{journal}{Rand coporation}  (\bibinfo{year}{1952})
  \bibinfo{pages}{15}.
%Type = Article
\bibitem[{Strumbelj and Kononenko(2013)}]{Strumbelj13}
\bibinfo{author}{E.~Strumbelj}, \bibinfo{author}{I.~Kononenko},
\newblock \bibinfo{title}{Explaining prediction models and individual
  predictions with feature contributions},
\newblock \bibinfo{journal}{Knowledge and Information Systems}
  \bibinfo{volume}{41} (\bibinfo{year}{2013}) \bibinfo{pages}{647--665}.
%Type = Inproceedings
\bibitem[{Miller Janny Ariza-Garz\'{o}n and Segovia-Vargas(2020)}]{Ariza20}
\bibinfo{author}{A.~C. Miller Janny Ariza-Garz\'{o}n, Javier~Arroyo},
  \bibinfo{author}{M.-J. Segovia-Vargas},
\newblock \bibinfo{title}{Explainability of a machine learning granting scoring
  model in peer-to-peer lending},
\newblock in: \bibinfo{booktitle}{Proceedings of IEEE Access},
  \bibinfo{year}{2020}.
%Type = Article
\bibitem[{Parsa et~al.(2020)Parsa, Movahedi, Taghipour, Derrible, and
  (Kouros)Mohammadian}]{Parsa20}
\bibinfo{author}{A.~B. Parsa}, \bibinfo{author}{A.~Movahedi},
  \bibinfo{author}{H.~Taghipour}, \bibinfo{author}{S.~Derrible},
  \bibinfo{author}{A.~(Kouros)Mohammadian},
\newblock \bibinfo{title}{Toward safer highways, application of xgboost and
  shap for real-time accident detection and feature analysis},
\newblock \bibinfo{journal}{Accident Analysis \& Prevention}
  \bibinfo{volume}{136} (\bibinfo{year}{2020}) \bibinfo{pages}{105405}.
%Type = Misc
\bibitem[{Wachter et~al.(2018)Wachter, Mittelstadt, and Russell}]{Wachter18}
\bibinfo{author}{S.~Wachter}, \bibinfo{author}{B.~Mittelstadt},
  \bibinfo{author}{C.~Russell}, \bibinfo{title}{Counterfactual explanations
  without opening the black box: Automated decisions and the gdpr},
  \bibinfo{year}{2018}.
%Type = Inproceedings
\bibitem[{Ramaravind K.~Mothilal(2020)}]{Mothila20}
\bibinfo{author}{C.~T. Ramaravind K.~Mothilal, Amit~Sharma},
\newblock \bibinfo{title}{Examples are not enough, learn to criticize!
  criticism for interpretability},
\newblock in: \bibinfo{booktitle}{Proceedings of the 2020 Conference on
  Fairness, Accountability, and TransparencyJanuary}, \bibinfo{year}{2020}.
%Type = Article
\bibitem[{Pearl(2019)}]{Pear19}
\bibinfo{author}{J.~Pearl},
\newblock \bibinfo{title}{The seven tools of causal inference, with reflections
  on machine learning},
\newblock \bibinfo{journal}{Communications of ACM} \bibinfo{volume}{62}
  (\bibinfo{year}{2019}) \bibinfo{pages}{7}.
%Type = Article
\bibitem[{Bottou et~al.(2013)Bottou, Peters, Qui{{\~n}}onero-Candela, Charles,
  Chickering, Portugaly, Ray, Simard, and Snelson}]{Bottou13}
\bibinfo{author}{L.~Bottou}, \bibinfo{author}{J.~Peters},
  \bibinfo{author}{J.~Qui{{\~n}}onero-Candela}, \bibinfo{author}{D.~X.
  Charles}, \bibinfo{author}{D.~M. Chickering}, \bibinfo{author}{E.~Portugaly},
  \bibinfo{author}{D.~Ray}, \bibinfo{author}{P.~Simard},
  \bibinfo{author}{E.~Snelson},
\newblock \bibinfo{title}{Counterfactual reasoning and learning systems: The
  example of computational advertising},
\newblock \bibinfo{journal}{Journal of Machine Learning Research}
  \bibinfo{volume}{14} (\bibinfo{year}{2013}) \bibinfo{pages}{3207--3260}.
%Type = Misc
\bibitem[{Johansson et~al.(2016)Johansson, Shalit, and
  Sontag}]{johansson2016learning}
\bibinfo{author}{F.~D. Johansson}, \bibinfo{author}{U.~Shalit},
  \bibinfo{author}{D.~Sontag}, \bibinfo{title}{Learning representations for
  counterfactual inference}, \bibinfo{year}{2016}.
%Type = Misc
\bibitem[{Schulam and Saria(2017)}]{schulam2017reliable}
\bibinfo{author}{P.~Schulam}, \bibinfo{author}{S.~Saria},
  \bibinfo{title}{Reliable decision support using counterfactual models},
  \bibinfo{year}{2017}.
%Type = Misc
\bibitem[{Neto(2020)}]{Neto20}
\bibinfo{author}{E.~C. Neto}, \bibinfo{title}{Towards causality-aware
  predictions in static machine learning tasks: the linear structural causal
  model case}, \bibinfo{year}{2020}.
%Type = Book
\bibitem[{Pearl(1988)}]{Pearl88}
\bibinfo{author}{J.~Pearl}, \bibinfo{title}{Probabilistic Reasoning in
  Intelligent Systems: Networks of Plausible Inference},
  \bibinfo{publisher}{Morgan Kaufmann Publishers}, \bibinfo{year}{1988}.
%Type = Book
\bibitem[{Russel and Norvig(2010)}]{russel10}
\bibinfo{author}{S.~Russel}, \bibinfo{author}{P.~Norvig},
  \bibinfo{title}{Artificial Intelligence: A Modern Approach},
  \bibinfo{publisher}{Pearson Education (3rd Edition)}, \bibinfo{year}{2010}.
%Type = Book
\bibitem[{Koller and Friedman(2009)}]{koller09prob}
\bibinfo{author}{D.~Koller}, \bibinfo{author}{N.~Friedman},
  \bibinfo{title}{Probabilistic Graphical Models: Principles and Techniques},
  \bibinfo{publisher}{The MIT Press}, \bibinfo{year}{2009}.
%Type = Article
\bibitem[{Scutari et~al.(2019)Scutari, Vitolo, and Tucker}]{Scutari19}
\bibinfo{author}{M.~Scutari}, \bibinfo{author}{C.~Vitolo},
  \bibinfo{author}{A.~Tucker},
\newblock \bibinfo{title}{Learning bayesian networks from big data with greedy
  search: computational complexity and efficient implementation},
\newblock \bibinfo{journal}{Statistics and Computing} \bibinfo{volume}{29}
  (\bibinfo{year}{2019}) \bibinfo{pages}{1095--1108}.
%Type = Article
\bibitem[{Heckerman et~al.(????)Heckerman, Geiger, and Maxwell}]{Heckerman95}
\bibinfo{author}{D.~Heckerman}, \bibinfo{author}{D.~Geiger},
  \bibinfo{author}{D.~Maxwell},
\newblock \bibinfo{title}{Learning bayesian networks: The combination of
  knowledge and statistical data},
\newblock \bibinfo{journal}{Machine Learning} \bibinfo{volume}{20} (\bibinfo{year}{1995})
  \bibinfo{pages}{197--243}.
%Type = Techreport
\bibitem[{Heckerman(1995)}]{Heckerman95Bayesian}
\bibinfo{author}{D.~Heckerman}, \bibinfo{title}{A Tutorial on Learning with
  Bayesian Networks}, \bibinfo{type}{Technical Report}, Microsoft Research
  Advanced Technology Division, Microsoft Corporation, \bibinfo{year}{1995}.
%Type = Techreport
\bibitem[{Chickering and Heckerman(1994)}]{Chickering94}
\bibinfo{author}{D.~M. Chickering}, \bibinfo{author}{D.~Heckerman},
  \bibinfo{title}{Learning Bayesian networks is NP-hard},
  \bibinfo{type}{Technical Report}, Tech. Rep. MSR-TR-94-17, Microsoft
  Corporation, \bibinfo{year}{1994}.
%Type = Inbook
\bibitem[{Chickering(1996)}]{Chickering96}
\bibinfo{author}{D.~M. Chickering}, \bibinfo{title}{Learning Bayesian Networks
  is NP-Complete}, \bibinfo{publisher}{Springer New York},
  \bibinfo{year}{1996}, pp. \bibinfo{pages}{121--130}.
%Type = Article
\bibitem[{Gabbay and Woods(2006)}]{Gabbay:Woods:2006}
\bibinfo{author}{D.~Gabbay}, \bibinfo{author}{J.~Woods},
\newblock \bibinfo{title}{Advice on abductive logic},
\newblock \bibinfo{journal}{Logic Journal of the IGPL} \bibinfo{volume}{14}
  (\bibinfo{year}{2006}) \bibinfo{pages}{189--219}.
%Type = Article
\bibitem[{Piri et~al.(2018)Piri, Delen, and Liu}]{Piri18}
\bibinfo{author}{S.~Piri}, \bibinfo{author}{D.~Delen},
  \bibinfo{author}{T.~Liu},
\newblock \bibinfo{title}{A synthetic informative minority over-sampling (simo)
  algorithm leveraging support vector machine to enhance learning from
  imbalanced datasets},
\newblock \bibinfo{journal}{Decision Support Systems} \bibinfo{volume}{106}
  (\bibinfo{year}{2018}) \bibinfo{pages}{15--29}.

\end{thebibliography}

\end{document}